\documentclass[10pt,twocolumn,letterpaper]{article}

\usepackage{iccv}
\usepackage{times}
\usepackage{epsfig}
\usepackage{graphicx}
\usepackage{amsmath}
\usepackage{amssymb}
\usepackage{amsthm}
\usepackage{tabularx}
\usepackage{multirow}
\usepackage{authblk}

\usepackage{textcomp,subfigure,multirow,upgreek}
\usepackage{booktabs}
\usepackage{mdwlist}

\usepackage{ragged2e}
\usepackage{bm}
\usepackage{fontawesome}
\usepackage{diagbox}
\usepackage{enumitem}  
\setenumerate[1]{itemsep=0pt,partopsep=0pt,parsep=\parskip,topsep=5pt}  
\setitemize[1]{itemsep=0pt,partopsep=0pt,parsep=\parskip,topsep=5pt}  
\setdescription{itemsep=0pt,partopsep=0pt,parsep=\parskip,topsep=5pt} 
\usepackage[pagebackref=true,breaklinks=true,letterpaper=true,colorlinks,bookmarks=false]{hyperref}

\iccvfinalcopy 

\def\iccvPaperID{7519} 

\ificcvfinal\fi
\usepackage{cleveref}
\newtheorem{thm}{Theorem}
\newtheorem{duplicate}{Theorem}

\begin{document}

\title{Why Approximate Matrix Square Root Outperforms Accurate SVD in Global Covariance Pooling?}
\author{Yue Song}
\author{Nicu Sebe}
\author{Wei Wang}
\affil{DISI, University of Trento, Trento, Italy \authorcr
{\tt\small \{yue.song nicu.sebe wei.wang\}@unitn.it}}
\setlength{\abovedisplayskip}{1pt}
\setlength{\belowdisplayskip}{1pt}

\maketitle
\ificcvfinal\thispagestyle{empty}\fi

\begin{abstract}
   Global Covariance Pooling (GCP) aims at exploiting the second-order statistics of the convolutional feature. Its effectiveness has been demonstrated in boosting the classification performance of Convolutional Neural Networks (CNNs). Singular Value Decomposition (SVD) is used in GCP to compute the matrix square root. However, the approximate matrix square root calculated using Newton-Schulz iteration~\cite{li2018towards} outperforms the accurate one computed via SVD~\cite{li2017second}. We empirically analyze the reason behind the performance gap from the perspectives of data precision and gradient smoothness. Various remedies for computing smooth SVD gradients are investigated. Based on our observation and analyses, a hybrid training protocol is proposed for SVD-based GCP meta-layers such that competitive performances can be achieved against Newton-Schulz iteration. Moreover, we propose a new GCP meta-layer that uses SVD in the forward pass, and Pad\'e approximants in the backward propagation to compute the gradients. The proposed meta-layer has been integrated into different CNN models and achieves state-of-the-art performances on both large-scale and fine-grained datasets. 
 
\end{abstract}

\vspace{-0.2cm}
\section{Introduction}
Global Covariance Pooling (GCP) explores the second-order statistics by normalizing the covariance matrix of the convolutional features before feeding them to the fully-connected layer. It has been shown to outperform the first-order pooling methods (\textit{e.g.,} max-pooling and average-pooling)~\cite{ionescu2015matrix,lin2015bilinear,li2017second,lin2017improved,li2018towards}. 
Generally, a GCP meta-layer computes the covariance matrix of the features as the global representation, and then performs eigendecomposition to derive the corresponding eigenvalues and eigenvectors, followed by normalization using either matrix logarithm~\cite{ionescu2015matrix,lin2015bilinear} or the matrix square root~\cite{li2017second,lin2017improved,li2018towards}. However, the logarithm may change the magnitude of eigenvalues significantly and over-stretch the small ones.
Besides, the matrix square root has been proved to amount to robust covariance estimation and to approximately exploit Riemannian geometry~\cite{li2017second}. Thus, the matrix square root is often preferred over logarithm normalization.

One can either use SVD to compute the accurate square root~\cite{li2017second} or use the Newton-Schulz iteration method to derive the approximate square root~\cite{li2018towards,lin2017improved}. Intuitively, the accurate one should yield better performance. Surprisingly, the approximate square root outperforms the exact one continuously~\cite{li2018towards}. Our paper starts with this intriguing observation and intends to find out the underlying reasons.

\begin{figure}[t]
\vspace{-0.4cm} 
   \begin{minipage}{.49\linewidth}
  \centering
  \centerline{\includegraphics[width=.99\linewidth]{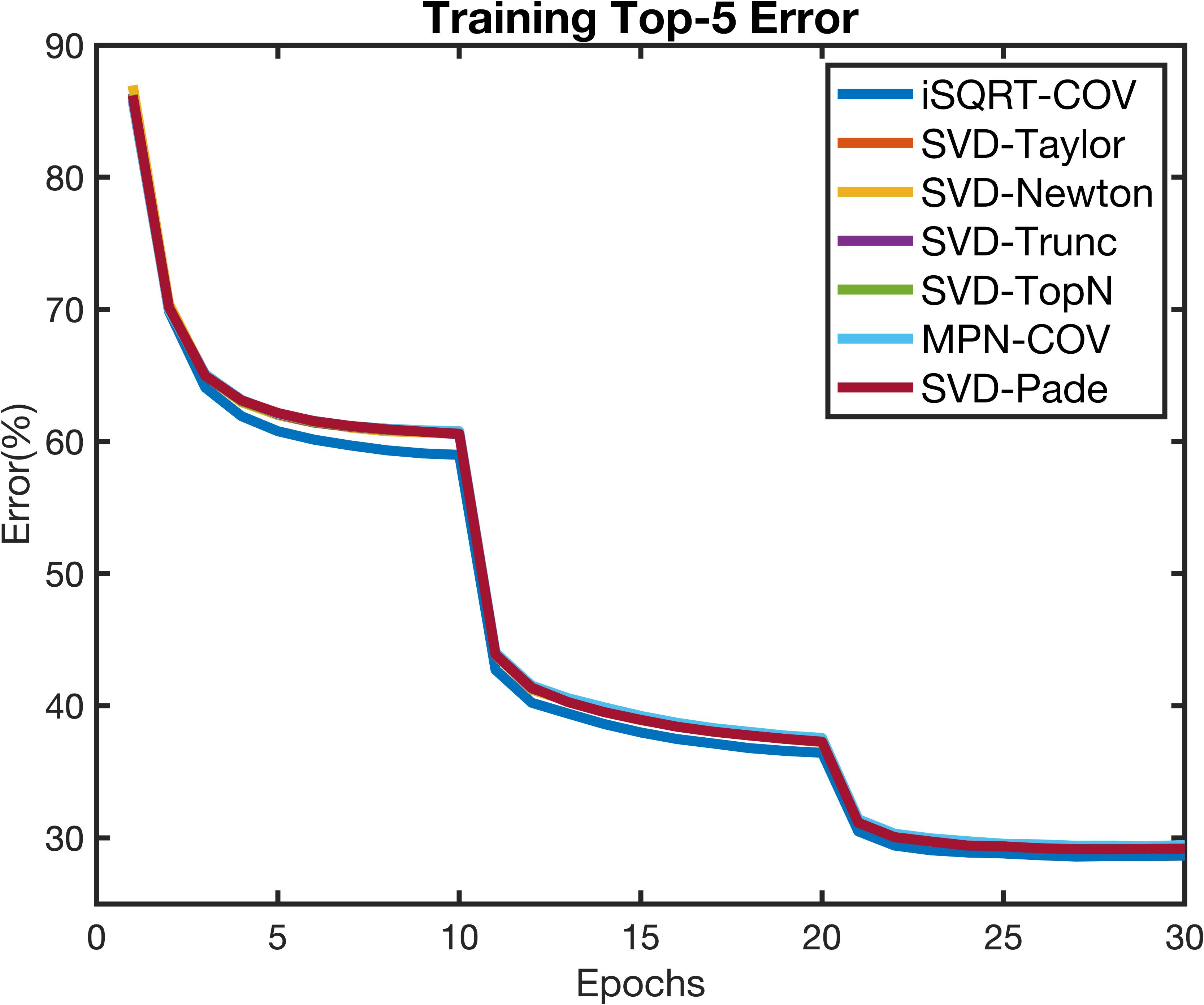}}
\end{minipage}
\hfill
\begin{minipage}{.49\linewidth}
  \centering
  \centerline{\includegraphics[width=.99\linewidth]{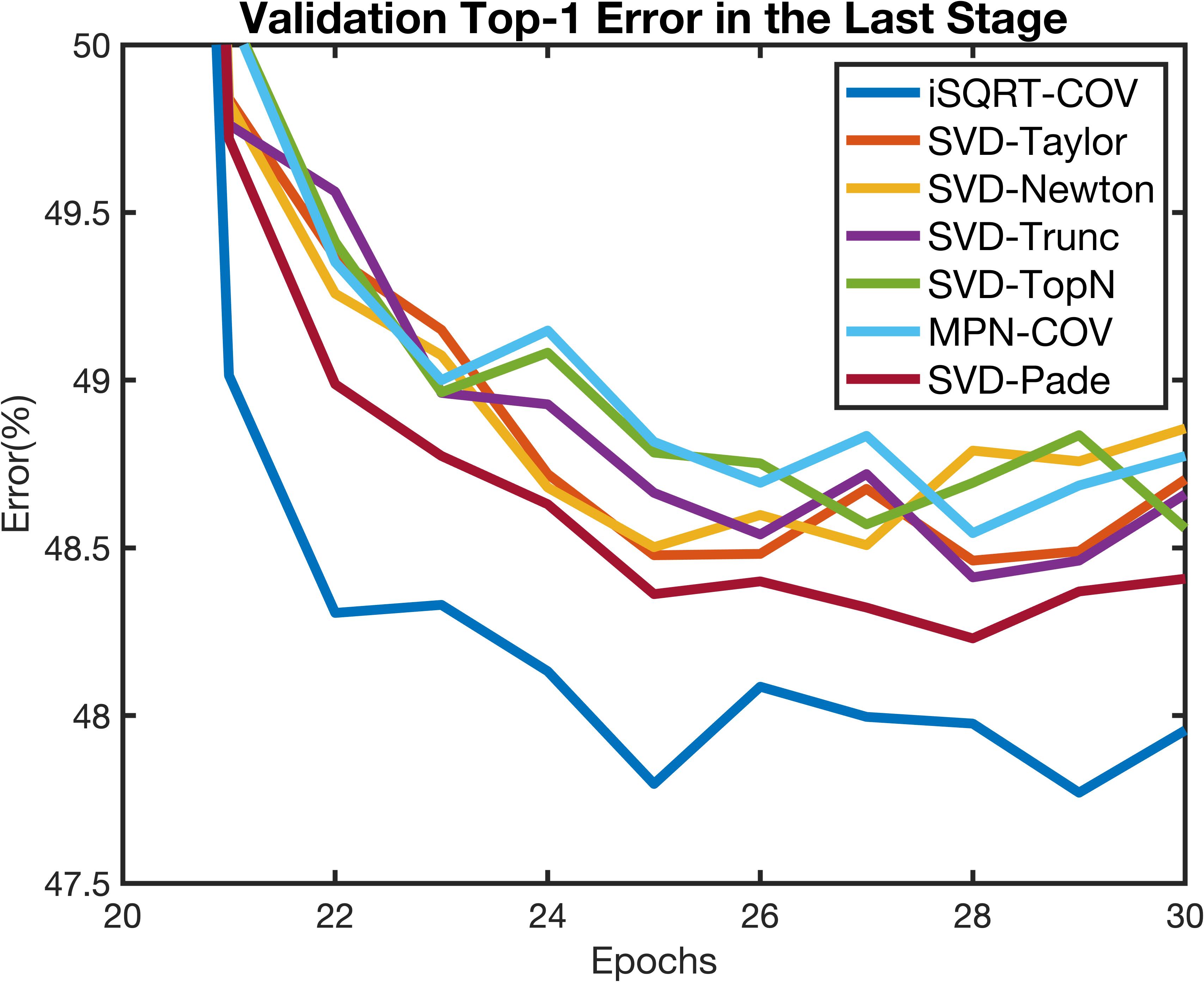}}
\end{minipage}
\caption{(\emph{Left}) The training top-5 error of AlexNet. The performance gap is gradually minimizing as the learning rate decays. (\emph{Right}) The validation top-1 error in the last stage. These SVD remedies marginally improve the performance but are not comparable against iSQRT-COV~\cite{li2018towards}. Our proposed SVD-Pad\'e achieves the best results among the SVD methods.}
\label{fig:cover}
\vspace{-0.4cm} 
\end{figure}

One crucial issue of SVD is the numerical instability of its gradients which is derived from the skew-symmetric matrix $\mathbf{K}$ whose off-diagonal elements are defined as $K_{ij}{=}1/(\lambda_{i}{-}\lambda_{j})$, where $\lambda_{i}$ and $\lambda_{j}$ are eigenvalues. $K_{ij}$ can be reformulated as $\frac{1}{\lambda_{i}} \frac{1}{1-{\lambda_{j}}{/}{\lambda_{i}}}$.
When the two eigenvalues are very small and close to each other, $\frac{1}{\lambda_{i}}$ and  $\frac{1}{1-{\lambda_{j}}{/}{\lambda_{i}}}$ will move towards infinity. As a consequence, the gradient $K_{ij}$ will explode. To avoid this issue, several attempts have been made to smooth the gradients~\cite{lin2017improved,huang2018decorrelated,wang2019backpropagation}. 
These methods consistently outperform ordinary SVD in other tasks, but none of them has been validated in GCP yet.

The gradient instability issue becomes more critical for GCP as it usually deals with very large matrices. According to our observation, when the covariance matrix dimension is very large (${>}200$), it is more likely to have many small eigenvalues. 
Single precision ({\ie, float32}) usually zeros out small eigenvalues and cannot guarantee the convergence of the network. Therefore, the data type of the covariance is set to double precision ({\ie, float64}) such that the small eigenvalues can be well represented. However, the high precision can also represent very subtle differences between the small eigenvalues. This can easily result in large gradients and aggravate the instability issue, and thus lead to inferior performance. Therefore, a couple of questions arise: 
\vspace{-0.2cm}
\begin{itemize}
\setlength{\itemsep}{0pt}
\setlength{\parsep}{0pt}
\setlength{\parskip}{0pt}
    \item[1)] Is the performance gap between MPN-COV~\cite{li2017second} (with accurate SVD) and iSQRT-COV~\cite{li2018towards} (with approximate one) related to their gradient smoothness?
    \item[2)] Can we smooth the gradient of the accurate SVD to help MPN-COV~\cite{li2017second} to achieve competitive performance against iSQRT-COV~\cite{li2018towards}?
\end{itemize}
\vspace{-0.2cm}
To answer these questions, we introduce several SVD backward remedies into MPN-COV~\cite{li2017second} which use different tricks (\emph{e.g.,} gradient truncation, abandoning small eigenvalues, and Taylor polynomial approximation~\cite{wang2019backpropagation}) to smooth the gradients. Fig.~\ref{fig:cover} shows the training and validation error curves of the modified SVD remedies and the ordinary SVD. We can see that although the modified SVD functions bring marginal performance gain over the ordinary SVD, still none of them is comparable against the Newton-Schulz based iSQRT-COV~\cite{li2018towards}. 
This implies that \emph{gradient smoothness does not fully account for the disparity}. 

Another interesting observation is that the performance gap between the modified SVD functions and iSQRT-COV~\cite{li2018towards} is gradually mitigating when the learning rate decreases (see Fig.~\ref{fig:cover} left). This is probably because with a small learning rate and stable network weights, the covariance matrices are more likely to be well-conditioned, and thus the smallest eigenvalues $\lambda_{min}$ are larger than \textsc{eps} ({\ie, the smallest positive number the data precision allows}). This could benefit SVD for stable eigendecompostion as smaller round-off errors and smoother gradients are obtained. Otherwise, if $\lambda_{min}$ is smaller than \textsc{eps}, the small eigenvalues would be zeroed out and the gradient $K_{ij}$ would go to infinity. We empirically 
show that the covariance matrices of MPN-COV~\cite{li2017second} are indeed becoming better-conditioned as the learning rate decreases, which is coherent with the trend of performance gap. This has enlightened us to combine the SVD function with iSQRT-COV~\cite{li2018towards} and to develop a hybrid training protocol, {\ie, use Newton-Schulz iteration to train the network until the learning rate is sufficiently small and the network weights are relatively stable, then switch to the ordinary/modified SVD for accurate matrix square root calculation}. By doing so, SVD only deals with well-conditioned matrices in the later stage. This hybrid strategy fully explores the potential of SVD for eigendecomposition, and thus these SVD methods achieve competitive and sometimes better performance than iSQRT-COV~\cite{li2018towards}. 

When the hybrid training protocol is used, unlike the situation in the standalone training, the marginal performance improvement of the modified SVD remedies over ordinary SVD no longer holds. The ordinary SVD function can sometimes outperform the modified SVD remedies with smooth gradients. This phenomenon makes us re-consider the effectiveness and necessity to smooth the gradients. As the large gradients can be close to infinity and are very likely to cause overflow, these modified SVD remedies drastically change the gradient at the order of magnitude for smoothness. However, the large but accurate gradients are important for learning robust representation and improving the generalization performance. We argue that \textbf{the large gradients should be closely approximated while avoiding singularities}. Among the SVD remedies, Wang~\textit{et al.}~\cite{wang2019backpropagation} suggested a promising direction to approximate the gradients using the Taylor polynomial, but their truncated Taylor series cannot converge under certain circumstances. Motivated by this work, we propose to use Pad\'e approximants, a rational approximation technique that has larger convergence radii and more powerful approximation abilities, to estimate the gradients. We show the appealing convergence property of Pad\'e approximants over Taylor polynomial. The proposed meta-layer outperforms all the existing spectral methods and its combination with Newton-Schulz iteration achieves state-of-the-art performances. Our contributions are threefold:
\vspace{-0.2cm}
\begin{itemize}[leftmargin=*]
    \item We empirically analyze the reason behind the superior performance of the approximate matrix square root over the accurate one from data precision and gradient smoothness view points. Various remedies for computing smooth SVD gradients are investigated and validated in GCP. 
    \item  A hybrid training protocol is proposed for GCP meta-layers and 
    competitive performance can be achieved compared with Newton-Schulz iteration. 
    We justified this strategy using the metric condition number to measure the ill-condition of covariance matrices.
    \item We propose a SVD backward algorithm that relies on Pad\'e approximants for fast and robust gradient approximation. It consistently achieves state-of-the-art performance on different datasets and different models. 
\end{itemize}
\vspace{-0.2cm}
Finally, to promote the easy applicability of the relevant SVD techniques, we will release the source codes of all the methods implemented in \textsc{Pytorch} upon acceptance\footnote{The code is available at \url{https://github.com/KingJamesSong/DifferentiableSVD}.}.
\section{Related Work}

\subsection{Differentiable SVD}
As a traditional matrix decomposition technique, SVD has a wide range of applications in modern deep learning, including batch whitening~\cite{wang2019backpropagation,huang2020investigation}, style transfer~\cite{chiu2019understanding,cho2019image}, and image segmentation~\cite{carreira2012semantic,ionescu2015matrix}. The theory of differentiable SVD was first formulated in~\cite{ionescu2015matrix,ionescu2015training}. When two eigenvalues are very close, the spurious gradient explosion is likely to happen and cause numerical instability. To circumvent this issue, \cite{huang2018decorrelated} proposed to divide the matrix into two smaller sub-matrices, which reduces the risk of having small and close eigenvalues. Nonetheless, this modification has no theoretical basis and may lead to inferior performances. 
\cite{wang2019backpropagation} proposed to rely on power iteration to iteratively calculate the approximate gradient. However, power iteration converges only if the largest eigenvalue $\lambda_{1}$ is dominant and this requirement may limit its practical usage (\ie, the top two largest values may be equal). \cite{wang2019backpropagation} proposed to use Taylor expansion for SVD gradient estimation. Due to the singularity of the function to approximate, the Taylor polynomial cannot give a good approximation when it is close to the polar singularity. Motivated by~\cite{wang2019backpropagation}, we propose to use Pad\'e approximants, a rational approximation technique to compute the SVD gradients. Compared with the Taylor polynomial, Pad\'e approximants have larger convergence radii and more powerful approximation abilities. 

\subsection{Global Covariance Pooling}
In deep neural networks, the global covariance pooling layer targets exploring the second-order statistics of the high-level representations before the fully-connected layer. Its powerful ability in boosting the network performance has been demonstrated in the past years~\cite{ionescu2015matrix,lin2015bilinear,wang2017g2denet,lin2017improved,li2017second,li2018towards,wang2019deep,wang2020deep,wang2020deepcvpr}. DeepO$^{2}$P~\cite{ionescu2015matrix} was the first end-to-end global covariance pooling network. It relies on SVD to compute the covariance and then conducts matrix logarithm for non-linear normalization. B-CNN~\cite{lin2015bilinear} was proposed to aggregate the outer product of convolutional features from two networks and then performs element-wise power normalization. Improved B-CNN~\cite{lin2017improved} investigated various ways of matrix normalization techniques and proved that the matrix square root normalization significantly outperforms other normalization schemes. Besides, they suggested alternate matrix square root computation techniques: using the Newton-Schulz iteration in the forward pass and solving the Lyapunov equation for gradient computation during backpropagation. G$^{2}$DeNet~\cite{wang2017g2denet} inserted a Gaussian distribution into the network and considered the geometry of the Gaussian manifold. MPN-COV~\cite{li2017second} presented a matrix power normalization method for robust covariance estimation. For a GPU-friendly computation concern, iSQRT-COV~\cite{li2018towards} proposed to use the Newton-Schulz iteration to accelerate the approximate matrix square root computation during both forward and backward calculations. Our paper conducts an investigation into the performance gap between~\cite{li2017second} and~\cite{li2018towards}. The former method computes the exact matrix square root, while the latter calculates the approximate one but achieves better performance.

\section{Investigation}


\subsection{Matrix Square Root Revisited}

Given the representation ${\mathbf{X}}{\in}{\mathbb{R}^{d{\times}N}}$ before the fully-connected layer of a neural network where $d$ defines the dimensionality and $N$ denotes the number of features, both accurate and approximate algorithms~\cite{li2017second,li2018towards} calculate the matrix square root of its sample covariance to exploit the second-order statistics. 

\noindent \textbf{Accurate Sqaure Root: MPN-COV~\cite{li2017second}.} 
After the convolutional layers, the extracted representation ${\mathbf{X}}{\in}{\mathbb{R}^{d\times N}}$ is used to compute the covariance matrix:
\begin{equation}
    \mathbf{P}=\mathbf{X}\Bar{\mathbf{I}}\mathbf{X}^{T}
    \label{covariance}
\end{equation}
where $\Bar{\mathbf{I}}=\frac{1}{N}(\mathbf{I}-\frac{1}{N}\mathbf{1}\mathbf{1}^{T})$ represents the centering matrix, $\mathbf{I}$ denotes the identity matrix, and $\mathbf{1}$ is a column vector whose values are all ones, respectively. After centralization, the covariance matrix $\mathbf{P}$ is symmetric positive semi-definite. The eigendecomposition can be performed via either SVD or EIG:
\begin{equation}
    \mathbf{P}=\mathbf{U}\mathbf{\Lambda}\mathbf{U}^{T}
    \label{SVD}
\end{equation}
here $\mathbf{\Lambda}=\rm diag(\lambda_{1},\dots,\lambda_{d})$ is the diagonal matrix with eigenvalues arranged in a non-increasing order, and $\mathbf{U}=[\mathbf{u}_{1},\dots,\mathbf{u}_{d}]$ is an orthogonal matrix where each column $\mathbf{u}_{i}$ is the eigenvector that corresponds to the eigenvalue $\lambda_{i}$. The matrix square root is obtained via the following equation:
\begin{equation}
    \mathbf{Q}\triangleq\mathbf{P}^{\frac{1}{2}}=\mathbf{U}\mathbf{F}(\mathbf{\Lambda}) \mathbf{U}^{T}
    \label{matrix_power}
\end{equation}
where $\mathbf{F}(\lambda)$ is the diagonal matrix $\rm diag(\lambda_{1}^{\frac{1}{2}},\dots,\lambda_{d}^{\frac{1}{2}})$. Afterwards, the global representation $\mathbf{Q}$ is fed into the fully-connected layer. 
During back-propagation, the partial derivative of loss function $l$  w.r.t. the input matrix $\mathbf{X}$ is computed based on the matrix backpropagation methodology~\cite{ionescu2015matrix,ionescu2015training}. Let $\frac{\partial l}{\partial \mathbf{Q}}$ denote the gradient from the last fully-connected layer, we can derive the partial derivative of eigenvector matrix $\mathbf{U}$ and eigenvalue matrix $\mathbf{\Lambda}$ as:
\begin{equation}
    \begin{gathered}
    \frac{\partial l}{\partial \mathbf{U}}=(\frac{\partial l}{\partial \mathbf{Q}} + (\frac{\partial l}{\partial \mathbf{Q}})^{T})\mathbf{U}\mathbf{F},\\
    \frac{\partial l}{\partial \mathbf{\Lambda}}=\frac{1}{2}(\rm{diag}(\lambda_{1}^{-\frac{1}{2}},\dots,\lambda_{d}^{-\frac{1}{2}})\mathbf{U}^{T} \frac{\partial \it{l}}{\partial \mathbf{Q}} \mathbf{U})_{\rm{diag}}
    \end{gathered}
    \label{SVD_de}
\end{equation}
where $(\cdot)_{\rm diag}$ denotes diagonalization that keeps only the diagonal elements of the matrix. Subsequently, the derivative of the covariance $\mathbf{P}$ can be calculated using the chain rule:
\begin{equation}
    \frac{\partial l}{\partial \mathbf{P}}=\mathbf{U}( (\mathbf{K}^{T}\circ(\mathbf{U}^{T}\frac{\partial l}{\partial \mathbf{U}}))+ (\frac{\partial l}{\partial \mathbf{\Lambda}})_{\rm diag})\mathbf{U}^{T}
    \label{COV_de}
\end{equation}
where $\circ$ represents matrix Hadamard product, and the matrix $\mathbf{K}$ is comprised of elements $K_{ij}{=}{1}/{(\lambda_{i}{-}\lambda_{j})}$ if $i{\neq}j$ and $K_{ij}{=}0$ otherwise. The loss $l$ w.r.t. the input feature $\mathbf{X}$ is finally computed as:
\begin{equation}
    \frac{\partial l}{\partial \mathbf{X}}=(\frac{\partial l}{\partial \mathbf{P}}+(\frac{\partial l}{\partial \mathbf{P}})^{T})\mathbf{X}\Bar{\mathbf{I}}
    \label{X_de}
\end{equation}
The whole network can be trained end-to-end using the forward and backward pass defined in~\cref{covariance,SVD,matrix_power,SVD_de,COV_de,X_de}. 
For the detailed loss derivations, the reader is kindly referred to~\cite{ionescu2015matrix,li2017second} for a comprehensive review.

\noindent \textbf{Approximate Square Root: iSQRT-COV~\cite{li2018towards}.} For the concern of GPU-friendly computation efficiency, \cite{li2018towards} proposed a loop-embedded meta-layer to iteratively compute the approximate matrix square root via Newton-Schulz iteration~\cite{higham2008functions}. Specifically, for computing the square root $\mathbf{Y}$ of a matrix $\mathbf{A}$, the coupled Newton-Schulz\ iteration takes the following form:
\begin{equation}
    \begin{gathered}
    \mathbf{Y}_{k}=\frac{1}{2}\mathbf{Y}_{k-1}(3\mathbf{I}-\mathbf{Z}_{k-1}\mathbf{Y}_{k-1}), \\
    \mathbf{Z}_{k}=\frac{1}{2}(3\mathbf{I}-\mathbf{Z}_{k-1}\mathbf{Y}_{k-1})\mathbf{Z}_{k-1}
    \end{gathered}
    \label{newton-schulz}
\end{equation}
where $\mathbf{Y}_{k}$ is initialized with $\mathbf{Y}_{0}{=}\mathbf{A}$, and $\mathbf{Z}_{k}$ starts with $\mathbf{Z}_{0}{=}\mathbf{I}$. Since Newton-Schulz iteration converges locally only if $\|\mathbf{A}{-}\mathbf{I}\|{<}1$, the covariance matrix $\mathbf{P}$ is first pre-normalized by its trace to meet the convergence condition:
\begin{equation}
    \mathbf{A}=\frac{1}{\rm{tr}(\mathbf{P})}\mathbf{P}
    \label{prenorm}
\end{equation}
Next, the normalized matrix $\mathbf{A}$ takes Newton-Schulz iteration and outputs the approximate matrix square root $\mathbf{Y}_{N}$ after $N$ iterations. The pre-normalization in~\cref{prenorm} non-trivially changes the data magnitude, which may cause the network to fail to converge. After Newton-Schulz iteration, the resultant matrix $\mathbf{Y}_{N}$ is post-compensated as follows:
\begin{equation}
    \mathbf{Q}=\sqrt{\rm{tr}(\mathbf{P})} \mathbf{Y}_{N}
\end{equation}

The back-propagation algorithm for iSQRT-COV~\cite{li2018towards} is not that straightforward as MPN-COV~\cite{li2017second}. The loss partial derivative $\frac{\partial l}{\partial \mathbf{Y}_{k}}$ and $\frac{\partial l}{\partial \mathbf{Z}_{k}}$ need to be calculated for each loop of Newton-Schulz iteration. For conciseness, we only give the derivative with the covariance here: 
\begin{equation}
\begin{split}
    \frac{\partial l}{\partial \mathbf{P}}=-\frac{1}{(\rm{tr}(\mathbf{P}))^2}\rm{tr}((\frac{\partial \it{l}}{\partial \mathbf{A}})^{\it{T}}\mathbf{P})\mathbf{I}+\frac{1}{\rm{tr}(\mathbf{P})}\frac{\partial \it{l}}{\partial \mathbf{A}}\\
    +\frac{1}{2\sqrt{\rm{tr}(\mathbf{P})}}\rm{tr}((\frac{\partial \it{l}}{\partial \mathbf{Q}})^{\it{T}}\mathbf{Y}_{\it{N}})\mathbf{I}
\end{split}
\end{equation}

Then~\cref{X_de} can be used again to derive the loss derivative with regards to the input feature.

\subsection{Data Precision Analysis}
\begin{figure}[t]
\vspace{-0.4cm}  
   \begin{minipage}{.49\linewidth}
  \centering
  \centerline{\includegraphics[width=.99\linewidth]{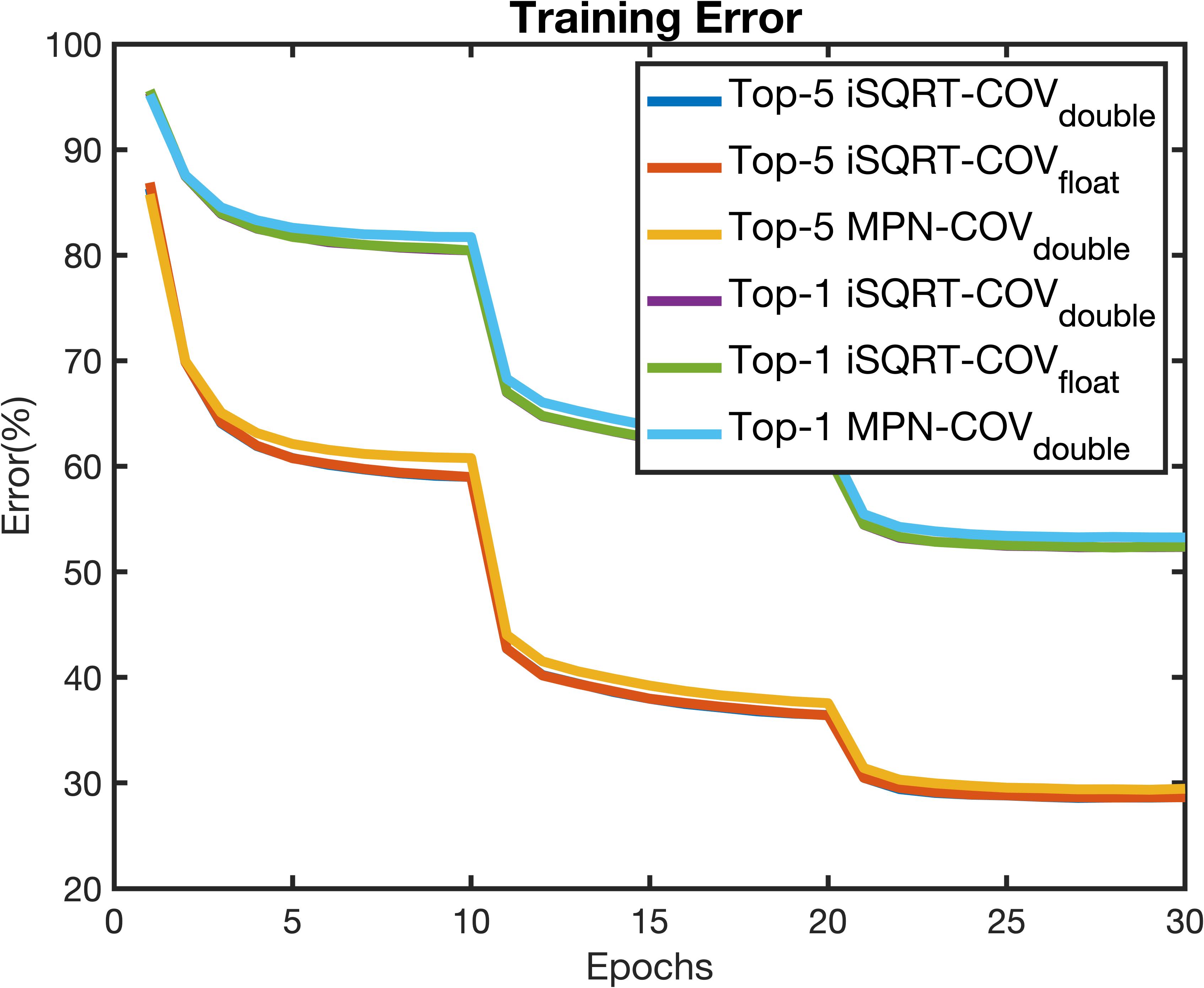}}
\end{minipage}
\hfill
\begin{minipage}{.49\linewidth}
  \centering
  \centerline{\includegraphics[width=.99\linewidth]{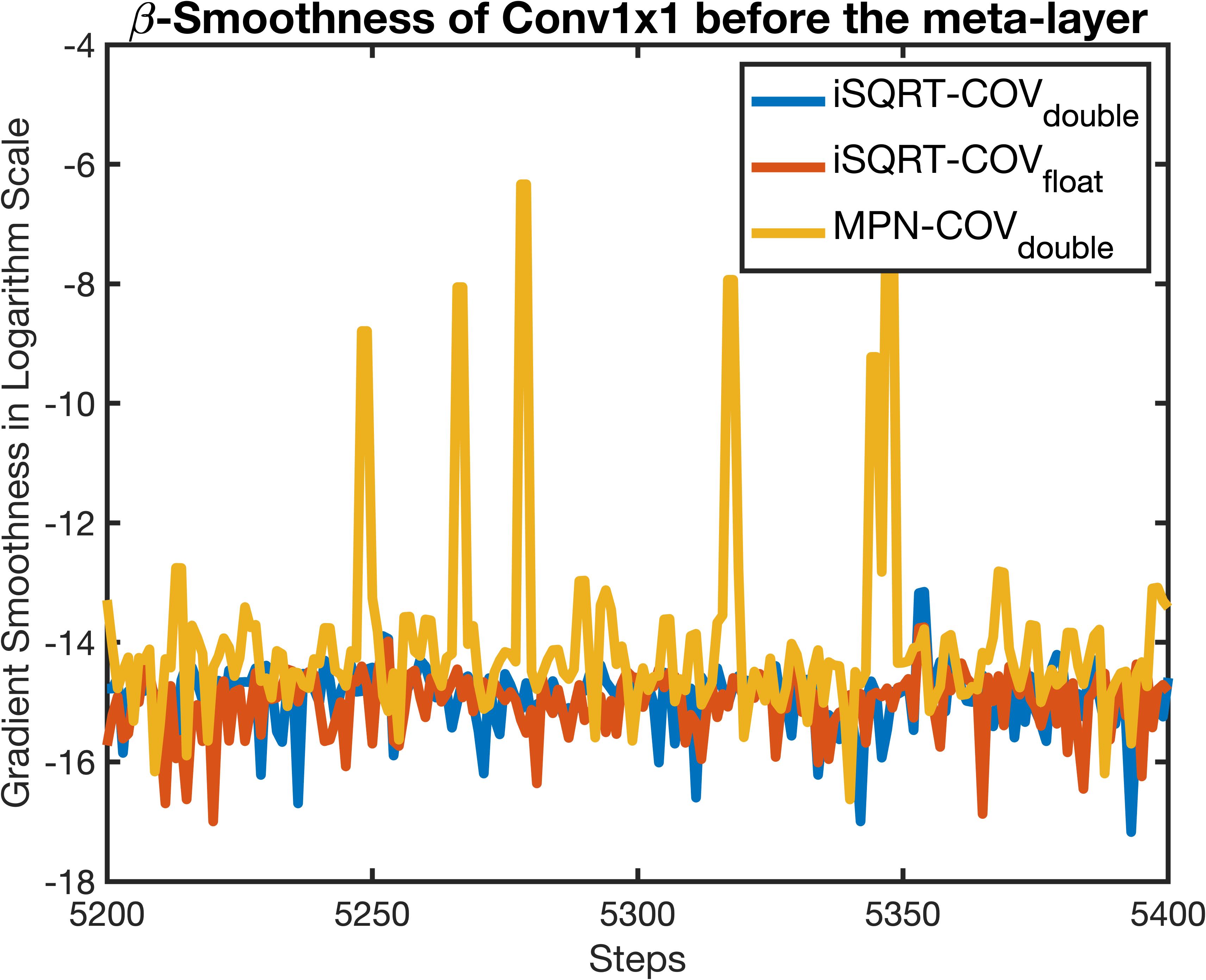}}
\end{minipage}
\caption{(\emph{Left}) Training error of both meta-layers in different precision. (\emph{Right}) The effective $\beta$-smoothness~\cite{nesterov2003introductory} of the two methods. Larger value indicates less smooth gradients.}
\vspace{-0.4cm}  
\label{fig:cov}
\end{figure}

As discussed before, GCP methods require double precision to ensure the effective numerical representation for small eigenvalues. To validate the impact of the data precision (\emph{i.e.,} single or double), we take AlexNet~\cite{krizhevsky2017imagenet} as the backbone and conduct experiments using both meta-layers in different data precision on ImageNet~\cite{deng2009imagenet}. AlexNet is chosen because it is light-weighted and easy to extrapolate to other deep models. The left of Fig.~\ref{fig:cov} presents the training error curve of the meta-layers. The data precision has a slight impact on iSQRT-COV~\cite{li2018towards} (about 0.1\%), whereas MPN-COV~\cite{li2017second} can be influenced substantially. MPN-COV~\cite{li2017second} can achieve reasonable performances in double precision but fails to converge using single precision. This observation confirmed the necessity of high precision to allow for effective numerical representation of eigenvalues. 

From~\cref{COV_de}, we can see that the small eigenvalues with high precision can easily result in large $K_{ij}$ and cause gradient overflow. We measure the effective $\beta$-smoothness~\cite{nesterov2003introductory} of both meta-layers to quantify their gradient smoothness. As can be observed in Fig.~\ref{fig:cov} right, MPN-COV~\cite{li2017second} has far less smooth gradients than iSQRT-COV~\cite{li2018towards}. 
To attain similar gradient smoothness for SVD, we explore different approaches to manipulate the backward algorithm of SVD for obtaining smooth gradients. In the rest of the paper, unless explicitly specified, we keep using double precision for GCP meta-layer of all the methods.

\subsection{Gradient Smoothness Analysis}
To investigate the concrete impact of gradient smoothness, we design several SVD meta-layers with smooth gradient. These meta-layers all use SVD as the forward pass but have different configurations during back-propagation. 

\noindent \textbf{Top-N Eigenvalue.}
The first approach is to directly abandon small eigenvalues that are likely to trigger numerical instability from the diagonal matrix $\mathbf{\Lambda}$. Since the diagonal eigenvalues are sorted in descending order, we can simply choose the top $N$ eigenvalues and discard the rest. The modification can be formally formulated as:
\begin{equation}
    \hat{\mathbf{\Lambda}}=\rm diag(\lambda_{1},\dots,\lambda_{\it N},0,\dots)
    \label{trunca_eps}
\end{equation}
The number of kept eigenvalues $N$ is chosen through cross-validation. This method is denoted as ``SVD-TopN".

\noindent \textbf{Gradient Truncation.}
Our second method is to limit the magnitude of gradient and apply truncation on matrix $\mathbf{K}$ during back-propagation. Specifically, we have 
\begin{equation}
    \hat{K}_{ij} = \begin{cases}
    \rm{T},\ \rm{if}\ \frac{1}{\lambda_{\it i}-\lambda_{\it j}}> T \\
    -\rm{T},\ \rm{if}\ \frac{1}{\lambda_{\it i}-\lambda_{\it j}}<-T
    \end{cases}
\end{equation}
where $\rm{T}$ is a large constant under the precision of the data type, and its optimal value can also be set by cross-validation. We name this approach ``SVD-Trunc".

\begin{figure}[t]
\vspace{-0.4cm}  
    \centering
    \includegraphics[width=0.9\linewidth]{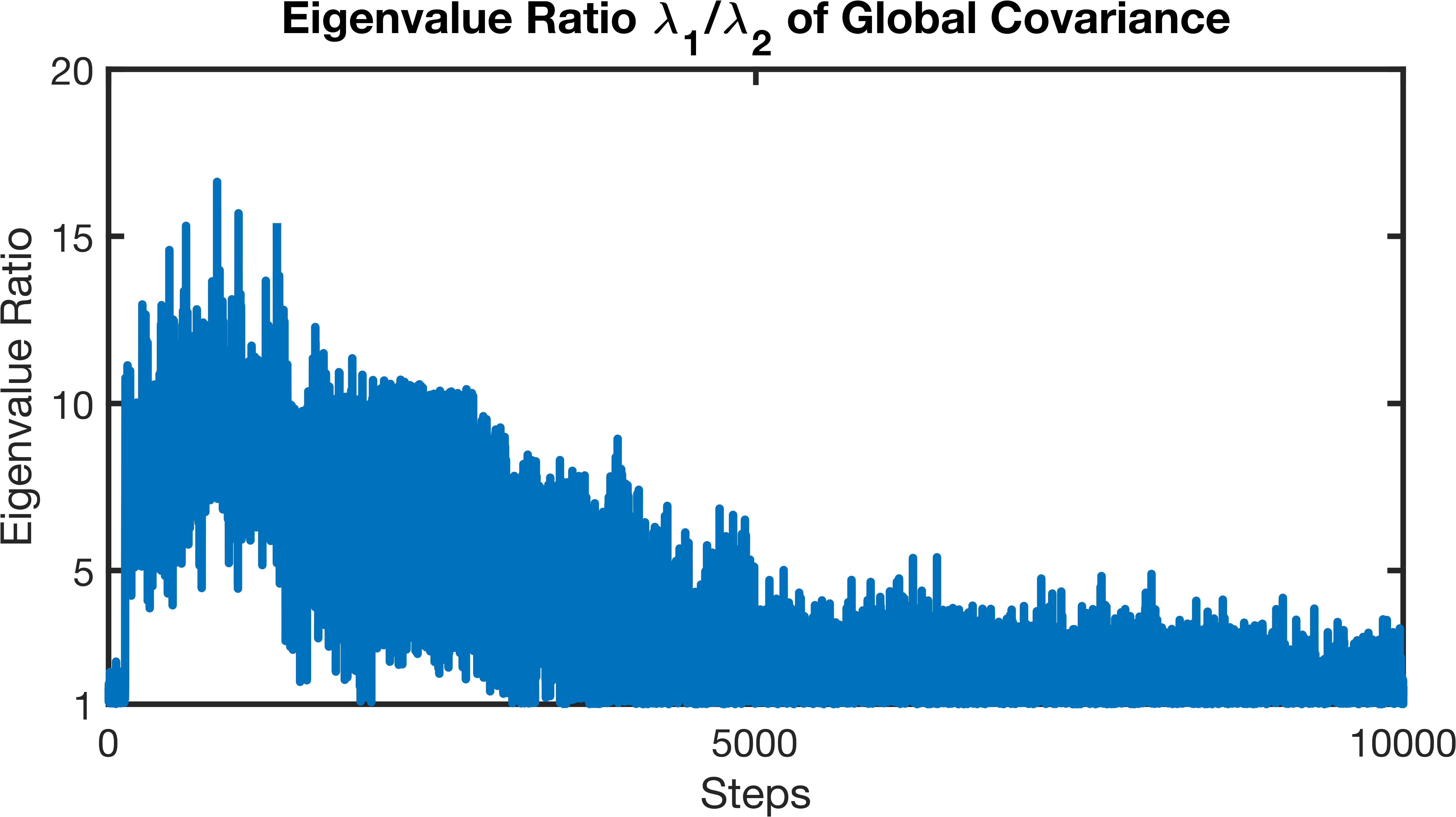}
    \caption{The ratio of the first two eigenvalues ($\lambda_{1}{/}\lambda_{2}$). The first eigenvalue is not dominant for every covariance matrix.}
    \label{fig:pi_ratio}
\vspace{-0.4cm} 
\end{figure}

\noindent \textbf{Power Iteration Gradient.} One recent SVD backward algorithm suggested using power iteration to compute the associated gradients~\cite{wang2019backpropagation}. Formally, power iteration takes the iterative update $\mathbf{u}^{k}{=}{\mathbf{P}\mathbf{u}^{k{-}1}}{/}{||\mathbf{P}\mathbf{u}^{k{-}1}||}$ to approximate the eigenvectors. By relying on $||\mathbf{P}\mathbf{u}_{1}||{=}||\mathbf{\lambda}_{1}\mathbf{u}_{1}||$ when $\lambda_{1}$ is dominant, the gradient in~\cref{COV_de} can be re-written as:
\begin{equation}
    \frac{\partial l}{\partial \mathbf{P}}=\mathbf{u}_{1}( (\mathbf{K}^{T}\circ(\mathbf{u}_{1}^{T}\frac{\partial l}{\partial \mathbf{u}_{1}}))+ (\frac{\partial l}{\partial \mathbf{\Lambda}})_{\rm diag})\mathbf{u}_{1}^{T}
\end{equation}
However, the convergence of power method requires that the first eigenvalue is dominant ($\lambda_{1}{/}\lambda_{2}{>}1$). We empirically observe that this method fails to converge in GCP. Fig.~\ref{fig:pi_ratio} displays the ratio of first two eigenvalues in the first $10,000$ training steps. As the network training goes, the ratio is gradually decreasing and reaches 1 for some covariance matrices. We thus conjecture that it cannot converge because the first eigenvalue $\lambda_{1}$ is not always dominant. 

\noindent \textbf{Newton-Schulz Gradient.} Our third remedy is to use Newton-Schulz iteration formulated in~\cref{newton-schulz} for back-propagation while using SVD as the forward pass. The iteration times are carefully tuned to achieve the best performances. This method is denoted as "SVD-Newton".

\noindent \textbf{Taylor Polynomial Gradient.} Recently, \cite{wang2019backpropagation} proposed to use Taylor expansion to approximate the SVD backward gradients. They re-formulated the non-zero elements of the matrix $\mathbf{K}$ computed in~\cref{COV_de} as a composition:
\begin{equation}
    K_{ij}=\frac{1}{\lambda_{i}-\lambda_{j}}=\frac{1}{\lambda_{i}}\cdot\frac{1}{1-(\lambda_{j}/\lambda_{i})}
    \label{K_reformulated}
\end{equation}
Notice that the right term resembles the function $f(x){=}1{/}(1{-}x)$. The Maclaurin series at $x{=}0$ of its Taylor expansion can be expressed as:
\begin{equation}
    P(z) = \sum\limits_{i=0}^{K}z^{i} + R(z^{K+1})
    \label{taylor}
\end{equation}
where $\sum_{n=0}^{K}z^{n}$ represents the Taylor expansion to degree $K$, and $R(z^{K+1})$ denotes the discarded remainder of higher degree. Injecting~\cref{taylor} into~\cref{K_reformulated} leads to:
\begin{equation}
    K_{ij}\approx\frac{1}{\lambda_{i}}(1+\frac{\lambda_{j}}{\lambda_{i}}+(\frac{\lambda_{j}}{\lambda_{i}})^2+\dots+(\frac{\lambda_{j}}{\lambda_{i}})^K)\leq\frac{K+1}{\lambda_{i}}
    \label{taylor_expansion}
\end{equation}
Now the numerical infinity $1{/}(\lambda_{i}-\lambda_{j})$ when the two eigenvalues are close has disappeared in the equation, and we have a bounded gradient estimation. From Cauchy root test, the Taylor series in~\cref{taylor} only converge in the range ${-1}{<}{z}{<}{1}$. In the case of ${j}{<}{i}$, the ratio ${\lambda_{j}}/{\lambda_{i}}$ is larger than $1$ and thus outsides the convergence radius. To avoid this issue, the matrix $\mathbf{K}$ is split into two triangular sub-matrix where the upper triangle defines the cases ${j}{>}{i}$ and the lower triangle consists of elements when ${j}{<}{i}$. As $\mathbf{K}$ is a skew-symmetric matrix, only the upper part needs to be computed. We call this method "SVD-Taylor".

\begin{figure}[t]
\vspace{-0.4cm} 
   \begin{minipage}{.49\linewidth}
  \centering
  \centerline{\includegraphics[width=.99\linewidth]{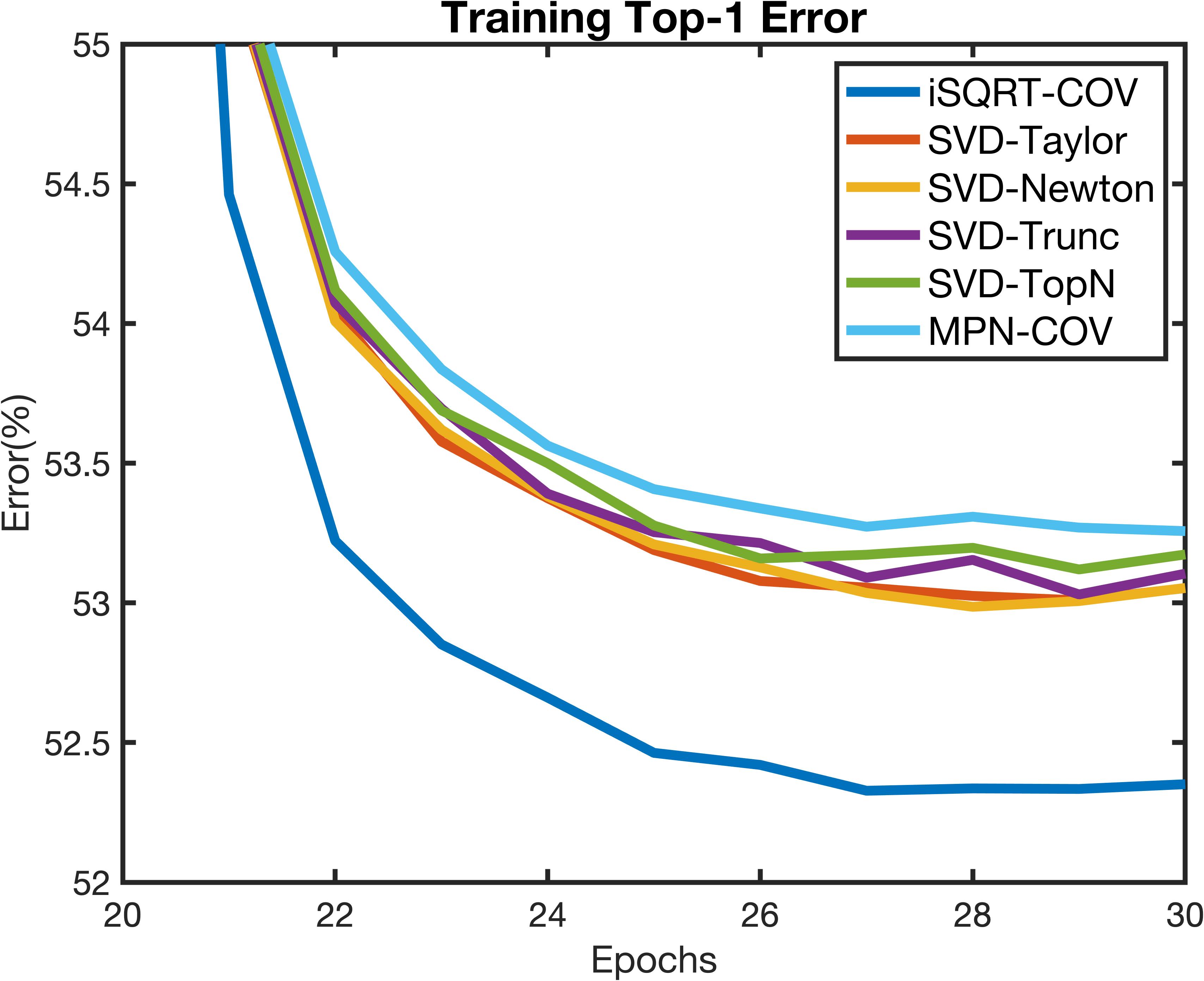}}
\end{minipage}
\hfill
\begin{minipage}{.49\linewidth}
  \centering
  \centerline{\includegraphics[width=.99\linewidth]{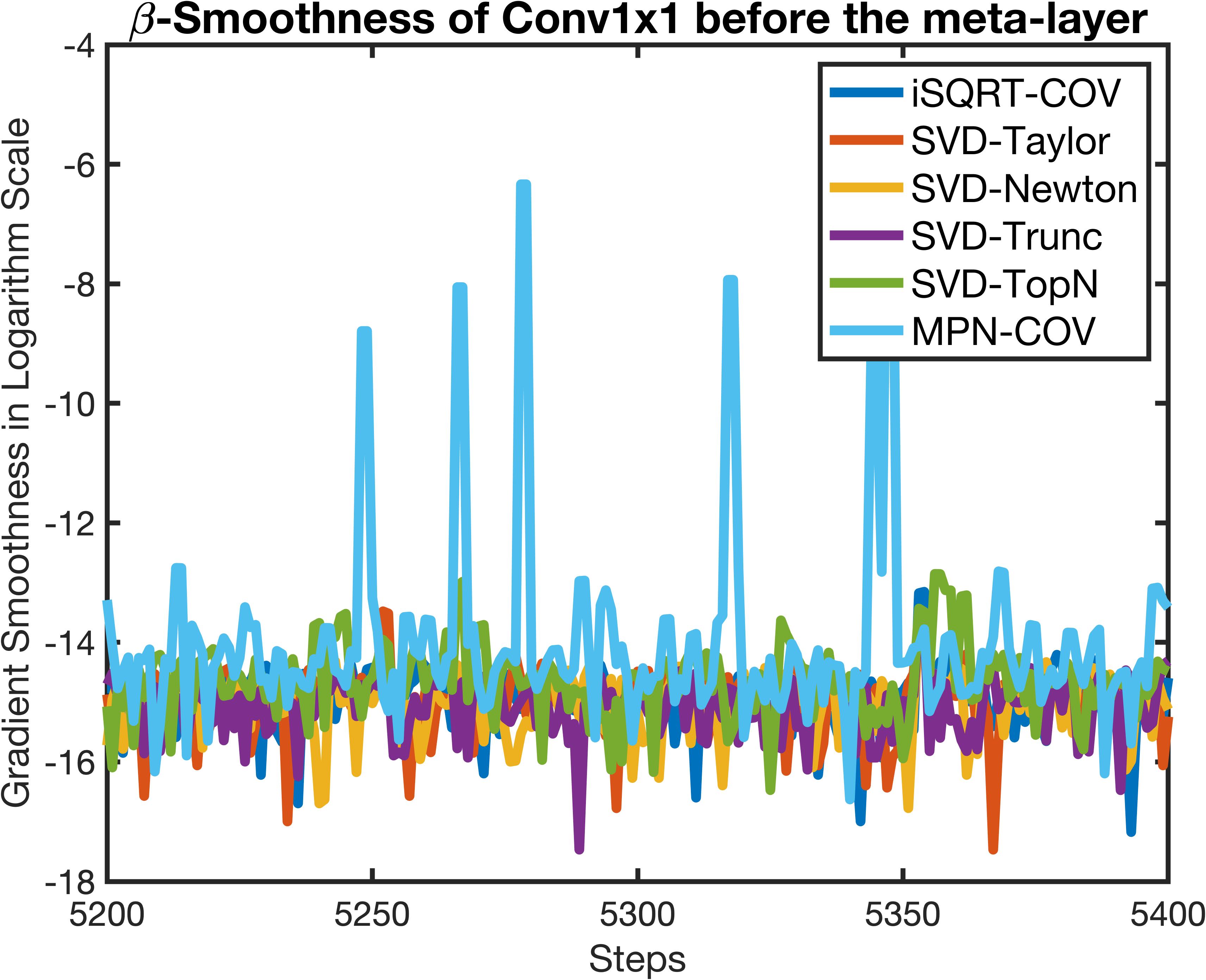}}
\end{minipage}
\caption{(\emph{Left}) Training top-1 error of our proposed meta-layers. These methods bring about maximally 0.2\% performance improvements over MPN-COV~\cite{li2017second}. (\emph{Right}) The effective $\beta$-smoothness~\cite{nesterov2003introductory} of the these methods. The gradient is smoothed to the similar degree with iSQRT-COV~\cite{li2018towards}.}
\label{fig:svd}
\vspace{-0.4cm} 
\end{figure}

We integrate the aforementioned methods into AlexNet~\cite{krizhevsky2017imagenet} and evaluate their performances on ImageNet~\cite{deng2009imagenet}. Fig.~\ref{fig:svd} compares the training top-1 error of these methods (left) and their effective $\beta$-smoothness~\cite{nesterov2003introductory} (right). The proposed SVD variants could smooth the SVD gradients to different extents and improve the performances by maximally 0.2\%. However, iSQRT-COV~\cite{li2018towards} still leads these SVD remedies by a large margin. 

\subsection{Hybrid Training Protocol}
Though these SVD remedies can not outperform iSQRT-COV~\cite{li2018towards}, their performance gap keeps decreasing as the learning rate decays. This phenomenon elicits our guess that the covariance matrices might be better-conditioned for SVD when the learning rate is sufficiently small and the model weights are well-trained. The hypothesis could be validated using condition number, {\ie, a metric to quantify the ill-condition of covariance matrices.} Condition number is defined by the ratio of the eigenvalues ${\lambda_{max}}{/}{\lambda_{min}}$ and can measure the stability of matrices. A high condition number indicates that $\lambda_{min}$ is relatively small and the matrix is close to singular, while a low condition number assures that $\lambda_{min}$ is relatively large and the matrix is well-conditioned. As a rule of thumb, when using double precision, matrices with condition numbers greater than $1e14$ are considered unstable and ill-conditioned. Ill-conditioned matrices are more likely to have eigenvalues that are smaller than \textsc{eps}. These small eigenvalues would be zeroed out and the gradient $K_{ij}$ might move to infinity, which could trigger round-off error and gradient explosion. We measure the average condition number of MPN-COV~\cite{li2017second} of AlexNet during different training stages. Table~\ref{tab:condition_number} shows the evaluation results of three epochs with different learning rates. As the training goes, the condition number gradually decreases. In the first stage, the average condition number is higher than the ill-condition threshold $1e14$. But in the later stages, MPN-COV~\cite{li2017second} shows much smaller average condition number and therefore has better-conditioned covariance matrices. This could benefit SVD for more stable eigendecomposition. 

\begin{table}[htbp]
    \centering
    \caption{Average condition number of different epochs. Higher value indicates the matrix is less stable and closer to singular.}
     \resizebox{0.7\linewidth}{!}{
    \begin{tabular}{r|c|c|c}
    \toprule
    Methods  & Epoch 1 & Epoch 11 & Epoch 21 \\
    \hline
    MPN-COV~\cite{li2017second} &3.41 e14 & 5.36 e13 &1.70 e13\\
    \bottomrule
    \end{tabular}
    }
    \label{tab:condition_number}
\vspace{-0.3cm} 
\end{table}

\begin{figure}[t]
   \begin{minipage}{.49\linewidth}
  \centering
  \centerline{\includegraphics[width=.99\linewidth]{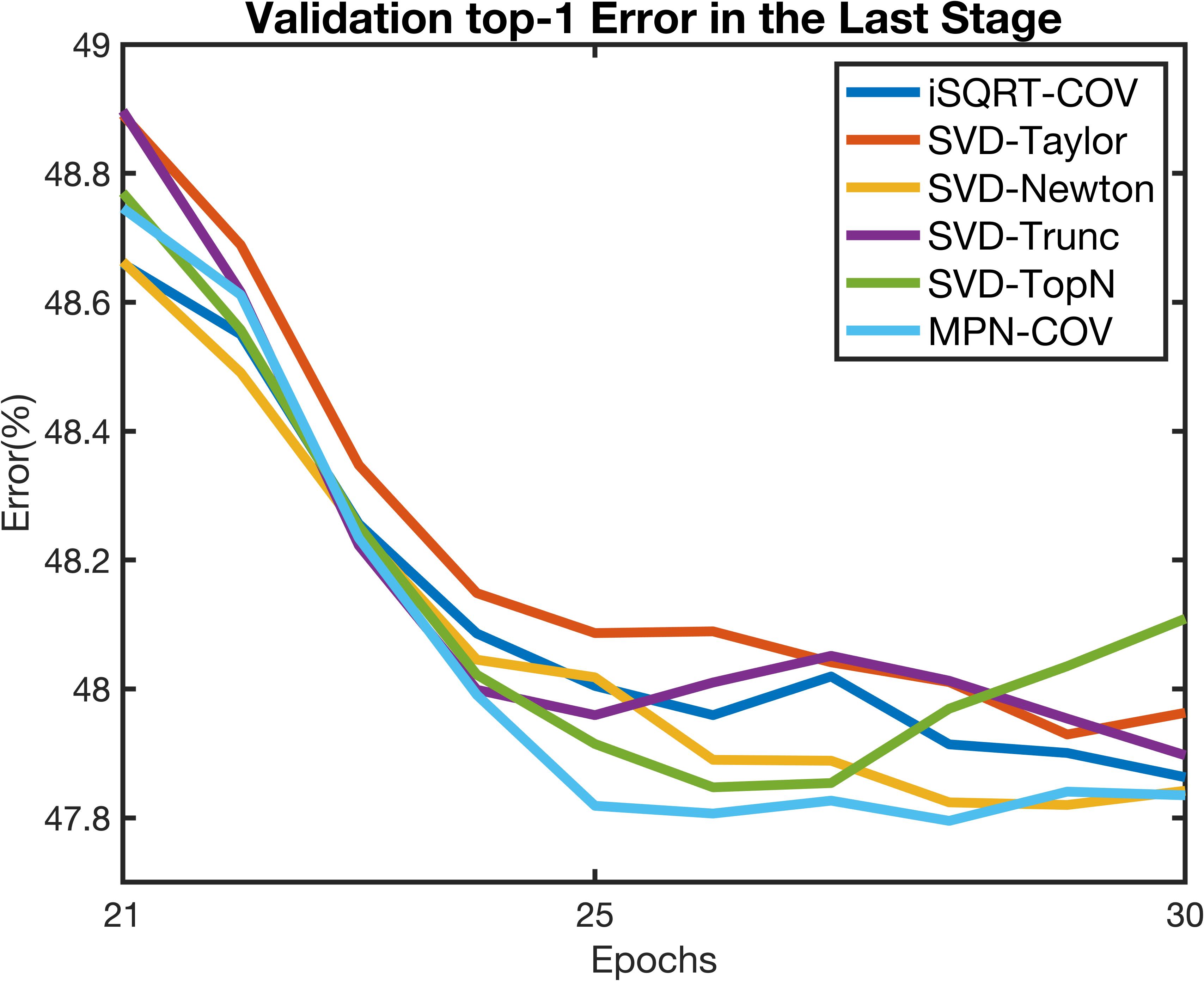}}
\end{minipage}
\hfill
\begin{minipage}{.49\linewidth}
  \centering
  \centerline{\includegraphics[width=.99\linewidth]{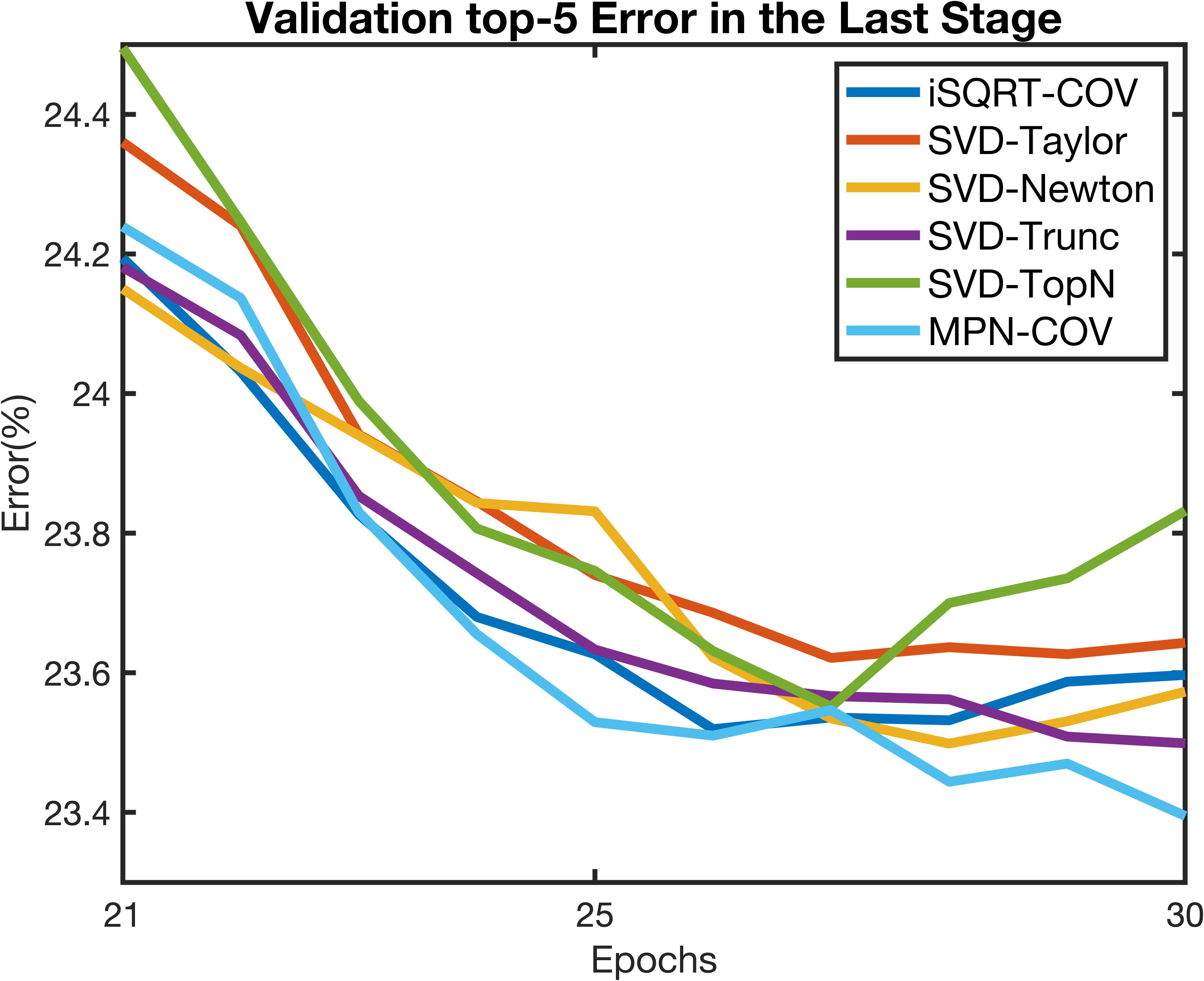}}
\end{minipage}
\caption{The validation top-1 error (left) and top-5 error (right) of AlexNet in the last stage using the hybrid training strategy. All the lines are smoothed by moving average filter for better view.}
\label{fig:warmup_error}
\vspace{-0.6cm} 
\end{figure}

This finding motivates us to design a hybrid training strategy such that SVD only needs to deal with well-conditioned matrices in later stages. In general, a deep model trained on ImageNet~\cite{deng2009imagenet} usually starts with a large learning rate and decays the rate twice when the error plateaus. We propose to use Newton-Schulz iteration to train the model before the last learning rate decay. Then the model is switched to SVD meta-layer to warm up the weights for few epochs. Finally, we decay the learning rate and keep using SVD meta-layer to tune the network till the end. Fig.~\ref{fig:warmup_error} shows the validation error of AlexNet using the proposed hybrid training strategy in the last stage. All the SVD variants have achieved competitive and even slightly better performances than iSQRT-COV~\cite{li2018towards}. Unlike the case of standalone training, when hybrid training strategy is applied, these SVD remedies with smooth gradients do not show obvious improvement over ordinary SVD. This observation challenges the necessity and effectiveness to smooth the gradients. 

\vspace{-2mm}
\section{Differentiable SVD by Pad\'e Approximants}
\vspace{-2mm}

Based on our above observation and analysis, we hazard that \textbf{the backward gradients should be closely approximated on the premise that the singularity is avoided}. However, existing methods sacrifice the large gradients for smoothness and none of them well approximate the gradients close to the singularities. Motivated by Taylor polynomial estimation~\cite{wang2019backpropagation}, we propose a more robust gradient approximation method for differentiable SVD.

\vspace{-2mm}
\subsection{Taylor Polynomial Problem.}
\vspace{-2mm}

\label{sec:taylor_problem}
Although Taylor polynomial provides a promising direction for gradient estimation, there still exist some caveats. First, $f(x){=}1{/}(1{-}x)$ is a meromorphic function and has a pole at $x{=}1$. Its corresponding geometric series $P(z){=}\sum_{i}^{\infty}z^{i}$ converge only when $|z|{<}1$. In the scope of approximation theory~\cite{powell1981approximation}, meromorphic functions that contain poles may not be well approximated by polynomial approximation techniques (\emph{e.g.}, Taylor and Chebyshev polynomial), as polynomials are entire functions without singularities. The closer the distance to the polar singularity ($x{=}1$) is, the worse its approximation ability would be. When the two eigenvalues are very close ($\lambda_{i}/\lambda_{j}{\approx}1$), the Taylor polynomial would move to the convergence boundary and reach the upper bound $K_{ij}{=}(K{+}1)/\lambda_{i}$. This would result in a poor approximation and make the Taylor polynomial sensitive to its truncation degree $K$. Rational approximation, on the other hand, usually gives a better approximation for function with poles because the use of rational functions allows singularities to be well represented. To address the aforementioned issues, we propose to substitute the truncated Taylor series with Pad\'e approximants, a kind of rational approximation that has enlarged convergence radii for more robust gradient estimation. In the next sections, we will introduce the definition of Pad\'e approximants, the formulation in the SVD algorithm, and their appealing convergence property over Taylor polynomial.


\vspace{-2mm}
\subsection{Pad\'e Approximants}
\vspace{-2mm}
The Pad\'e approximants are a particular type of rational fraction approximation. The idea is use the ratio of two polynomials to match a power series. Consider a Maclaurin Taylor series $A(x){=}{\sum_{i=0}^{\infty}}a_{i}x^{i}$ to match, we have the corresponding formal definition of Pad\'e approximants:
\begin{equation}
    [M/N]=\frac{P_{M}(x)}{Q_{N}(x)}=\frac{\sum\limits_{m=0}^{M}p_{m}x^{m}}{1+\sum\limits_{n=1}^{N}q_{n}x^{n}}
    \label{pade}
\end{equation}
where the numerator ${P_{M}(x)}$ is a polynomial of degree at most $M$, and the denominator ${Q_{N}(x)}$ is a polynomial of degree at most $N$. A ${[}{M}{/}{N}{]}$ Pad\'e approximants agree up to a Taylor series of order ${M}{+}{N}{+}{1}$. Their coefficients are determined by matching the power series:
\begin{equation}
    A(x)-\frac{P_{M}(x)}{Q_{N}(x)}=R(x^{M+N+1})
\end{equation}
By introducing~\cref{pade} into this equation and setting the remainder $R(x^{M{+}N{+}1})$ to zero, we can linearize the coefficient equations as:
\begin{equation}
    \begin{gathered}
    a_{0}=p_{0}\\
    a_{1}+a_{0}q_{1}=p_{1}\\
    \vdots\\
    a_{M}+a_{M-1}q_{1}+\dots+a_{0}q_{M}=p_{M}\\
    a_{M+1}+a_{M}q_{1}+\dots+a_{M-N+1}q_{N}=0\\
    \vdots\\
    a_{M+N+1}+a_{M+N-1}q_{1}+\dots+a_{M}q_{N}=0
    \end{gathered}
    \label{linear_solution}
\end{equation}
where $a_{n}{\equiv}0$ for $n{<}0$ and $q_{j}{\equiv}0$ for $j{>}M$. Solving these equations directly gives the Pad\'e coefficients.  

\vspace{-2mm}
\subsection{Formulation in SVD Back-propagation}
\vspace{-2mm}
Let us start with the gradient approximated by Taylor polynomial. The core idea of Taylor gradient is to use a truncated Taylor series in~\cref{taylor} of degree $K$ to approximate the elements of matrix $\mathbf{K}$ defined in~\cref{K_reformulated}. The Pad\'e approximants can be derived by matching~\cref{taylor}:
\begin{equation}
    \frac{P_{M}(x)}{Q_{N}(x)}+R(x^{M+N+1})=\sum\limits_{i=0}^{K}x^{i} + R(x^{K+1})
\end{equation}
Then we can get the matched Pad\'e approximants and re-express the non-zero matrix elements of $\mathbf{K}$ as
\begin{equation}
    K_{ij}\approx\frac{1}{\lambda_{i}}\cdot\frac{P_{M}({\lambda_{j}}/{\lambda_{i}})}{Q_{N}({\lambda_{j}}/{\lambda_{i}})}=\frac{1}{\lambda_{i}}\cdot\frac{\sum\limits_{m=0}^{M}p_{m}(\frac{\lambda_{j}}{\lambda_{i}})^{m}}{1+\sum\limits_{n=1}^{N}q_{n}(\frac{\lambda_{j}}{\lambda_{i}})^{n}}
    \label{pade_K}
\end{equation}
In practice, diagonal $[M/N]$ Pad\'e approximants where the numerator and denominator have the same degree ($M{=}N{+}1$) are often preferred. This will guarantee stable Pad\'e coefficients and the enlarged convergence range. 

\subsection{Enlarged Convergence Range.} 
Pad\'e approximants are derived out of a finite Taylor series, but they have more desired convergence properties than truncated Taylor series. Specifically for diagonal Pad\'e approximants, we have:
\begin{thm}
\vspace{-0.15cm}
If the function $f(z)$ is a Stieltjes transform $f(z){=}\int_{a}^{b}\frac{1}{z-x}d\mu(x)$ of a compactly supported measure $\mu(x)$ in $[a{,}b]$, then the associated $[N{+}1{/}N]$ diagonal Pad\'e approximants are orthogonal and there exists such function $r(x){>}1$ that the convergence $\lim\limits_{N\rightarrow\infty}|f(z){-}\frac{P_{N{+}1}(z)}{Q_{N}(z)}|^{\frac{1}{N}}{=}\frac{1}{r^2}$ is exponential in $[a{,}b]$. 
\vspace{-0.15cm}
\end{thm}
The convergence theorem along with the asymptotic behavior of diagonal Pad\'e approximants were given in~\cite{van2006pade}. Notice that the function $f(x){=}1{/}(1{-}x)$ is compactly supported in $\mathbb{R}$, as it only vanishes at infinity. Thus, this property can ensure that the associated Pad\'e approximants still have good approximations even close to the convergence boundary of the original Taylor series. That being said, a diagonal Pad\'e approximant can not only avoid singularities of the SVD gradients but also give a very close approximation for any possible eigenvalue ratio $\lambda_{j}{/}\lambda_{i}$. 
\section{Experiment}

\subsection{Models and Datasets}
Following~\cite{li2017second,li2018towards}, we first take AlexNet~\cite{krizhevsky2017imagenet} and ResNet-50~\cite{he2016deep} as the backbones and conduct experiments on ImageNet 2012~\cite{deng2009imagenet} for the large-scale visual recognition.  After training GCP models on ImageNet, we evaluate the methods on Fine-Grained Visual Categorization (FGVC). The pre-trained ResNet-50 with different GCP meta-layers using hybrid training strategy are  fine-tuned on three popular fine-grained benchmarks, \emph{i.e.,} Caltech Birds (Birds)~\cite{WelinderEtal2010}, Stanford Dogs (Dogs)~\cite{KhoslaYaoJayadevaprakashFeiFei_FGVC2011}, and Stanford Cars (Cars)~\cite{KrauseStarkDengFei-Fei_3DRR2013}. More details about the datasets and implementation can be found in the Appendix.

\subsection{Results on ImageNet with AlexNet}


\begin{table}[t]
\vspace{-0.3cm} 
    \centering
    \caption{Validation error of AlexNet using different training strategy. The best three results are highlighted in \textcolor{red}{red}, \textcolor{blue}{blue}, and \textcolor{green}{green}.}
     \resizebox{0.9\linewidth}{!}{
    \begin{tabular}{r|cc|cc|cc}
    \toprule
    \multirow{3}*{Methods} & \multicolumn{2}{c|}{Standalone Training} & \multicolumn{4}{c}{Hybrid Training}\\
    \cline{2-7}
    & \multicolumn{2}{c|}{Final Error (\%)} & \multicolumn{2}{c|}{Final Error (\%)} & \multicolumn{2}{c}{Best Error (\%)}\\
    \cline{2-7} 
    &top-1 & top-5 & top-1 & top-5 & top-1 & top-5\\
    \hline
    iSQRT-COV~\cite{li2018towards} &\textcolor{red}{47.95} & \textcolor{red}{23.64} &47.95 & 23.64 & 47.81 & 23.54 \\
    SVD-Pad\'e                       &\textcolor{blue}{48.41} & \textcolor{blue}{23.91}&\textcolor{red}{47.76} & \textcolor{blue}{23.48}& \textcolor{red}{47.63}&\textcolor{red}{23.21}\\
    SVD-Taylor                     &48.70 & 24.30 &\textcolor{green}{47.92} & 23.56&47.86 & 23.56\\
    SVD-Newton                     &48.86 & 25.08 &\textcolor{blue}{47.87} & \textcolor{blue}{23.48}&\textcolor{green}{47.77} & \textcolor{green}{23.38}\\
    SVD-Trunc                      &48.66 & \textcolor{green}{24.10}&47.98 & \textcolor{red}{23.45}&47.81 & 23.48\\
    SVD-TopN                       &\textcolor{green}{48.56} & \textcolor{green}{24.10}&47.96 & 23.65&47.81 & 23.54\\
    MPN-COV~\cite{li2017second}    &48.77 & 24.28& 47.94 & \textcolor{green}{23.54}&\textcolor{blue}{47.75} & \textcolor{blue}{23.24}\\
    \bottomrule
    \end{tabular}
    }
    \label{tab:alexnet_top15}
\vspace{-0.5cm} 
\end{table}

Table~\ref{tab:alexnet_top15} displays the validation errors of AlexNet using the standalone and hybrid training strategies. 
In the case of standalone training, the proposed SVD-Pad\'e method outperforms other SVD remedies significantly and improves the ordinary SVD algorithm by about 0.4\% in both top-1 and top-5 error. Since the performance gap between each method is subtle and may fluctuate when using the hybrid training strategy, we report both the final and best validation error for comprehensive comparisons. When hybrid training protocol is used, our proposed SVD-Pad\'e achieves state-of-the-art performances in four metrics and surpasses all the other baselines including iSQRT-COV~\cite{li2018towards}. 


\subsection{Results on ImageNet with ResNet}

\begin{table}[htbp]
\vspace{-0.4cm} 
    \centering
    \caption{Validation errors of ResNet-50 and ResNet-101. The best three results are highlighted in \textcolor{red}{red}, \textcolor{blue}{blue}, and \textcolor{green}{green}.}
     \resizebox{0.9\linewidth}{!}{
    \begin{tabular}{c|c|cc|cc}
    \toprule
     &\multirow{2}*{Methods} & \multicolumn{2}{c|}{Standalone Training} & \multicolumn{2}{c}{Hybrid Training}\\
    \cline{3-6} 
    & & top-1 & top-5 & top-1 & top-5 \\
    \hline
    \multirow{8}{*}{\rotatebox[origin=c]{90}{ResNet-50}}&iSQRT-COV [14] &\textcolor{blue}{{22.81}} & \textcolor{blue}{{6.60}} &22.81 &6.60  \\
    &SVD-Pad\'e                     &\textcolor{red}{{22.67}} & \textcolor{red}{{6.51}} &\textcolor{red}{{22.60}} &\textcolor{red}{{6.44}}\\
    &SVD-Taylor                     &22.91 &6.67 &22.77 &\textcolor{green}{6.53} \\
    &SVD-Newton                     &22.86 &\textcolor{green}{6.65} &\textcolor{blue}{22.72} &6.55 \\
    &SVD-Trunc                      &\textcolor{green}{{22.85}} &6.70 &\textcolor{green}{22.74} &6.52\\
    &SVD-TopN                       &22.91 &{6.68} &22.76 &{6.51}\\
    &MPN-COV [15]   &22.93 &6.75 &22.79 &\textcolor{blue}{6.50}\\
    \cline{2-6} 
    &Vanilla ResNet-50 &23.85 &7.13 &23.85 &7.13\\
    \hline
    \multirow{4}{*}{\rotatebox[origin=c]{90}{ResNet-101}}&iSQRT-COV [14] &21.60 & 5.88 & 21.60 & 5.88 \\
    & SVD-Pad\'e & \textcolor{red}{{21.48}} & \textcolor{red}{{5.80}} & \textcolor{red}{{21.40}} & \textcolor{red}{{5.69}}\\
    & MPN-COV [15]   &21.79 &5.99 & 21.58 &5.80\\
    \cline{2-6}
    &Vanilla ResNet-101 &22.63 &6.44  &22.63 &6.44\\
    \bottomrule
    \end{tabular}
    }
    \label{tab:resnet_top15}
    \vspace{-0.4cm} 
\end{table}


Tables~\ref{tab:resnet_top15} compares the validation errors of ResNet using different training strategies. The results are very coherent with those of AlexNet. Our SVD-Pad\'e achieves the best evaluation results. Even when the standalone training protocol is applied, \emph{i.e.,} the SVD methods are trained from scratch, the SVD-Pad\'e can still outperform~\cite{li2018towards} by about 0.2 \%. This demonstrates the necessity of closely approximating gradients. When it comes to the hybrid training strategy, these SVD variants have achieved competitive performances against iSQRT-COV~\cite{li2018towards} and even outperform by a narrow margin. In particular, our SVD-Pad\'e remedy also achieves the best results. We also believe that warming up more epochs when switching to SVD methods can a bring larger performance gain.


\subsection{Results on FGVC with ResNet}

\begin{table}[t]
    \centering
    \caption{Comparison of accuracy (\%) on fine-grained datasets using ResNet-50 with different GCP meta-layers as backbone. The best three results are highlighted in \textcolor{red}{red}, \textcolor{blue}{blue}, and \textcolor{green}{green} respectively. $/$ means the method cannot converge.}
     \resizebox{0.9\linewidth}{!}{
    \begin{tabular}{r|c|c|c|c|c|c}
    \toprule
    \multirow{2}*{Methods} & \multicolumn{2}{c|}{Birds~\cite{WelinderEtal2010}} & \multicolumn{2}{c|}{Dogs~\cite{KhoslaYaoJayadevaprakashFeiFei_FGVC2011}} & \multicolumn{2}{c}{Cars~\cite{KrauseStarkDengFei-Fei_3DRR2013}} \\
    \cline{2-7}
    & Final & Best & Final & Best & Final & Best \\
    \hline
    iSQRT-COV~\cite{li2018towards} &85.95 &86.45 &82.34 &83.45 &90.93 &91.56 \\
    SVD-Pad\'e                     &\textcolor{blue}{87.05} &\textcolor{red}{87.29} &\textcolor{red}{83.40} &\textcolor{red}{84.34} &\textcolor{red}{92.55} &\textcolor{blue}{92.99}\\ 
    SVD-Taylor                     &86.95 &87.20 &\textcolor{green}{83.23} &\textcolor{green}{84.22} &\textcolor{blue}{92.46} &92.71\\
    SVD-Newton                     &\textcolor{green}{86.97} &\textcolor{green}{87.22} &83.08 &83.94 &92.35 &92.51\\
    SVD-Trunc                      &\textcolor{red}{87.16} &\textcolor{blue}{87.25} &82.95 &84.03 &{92.43} &\textcolor{red}{93.04}\\
    SVD-TopN                       &/ &/ &/ &/ &/ &/\\
    MPN-COV~\cite{li2017second}    &86.89 &87.19 &\textcolor{blue}{83.34} &\textcolor{blue}{84.24} & \textcolor{green}{92.45} &\textcolor{green}{92.84}\\
    \bottomrule
    \end{tabular}
    }
    \label{tab:fgvc_acc}
    \vspace{-0.6cm} 
\end{table}

We finally validate the performances of these GCP meta-layers on FGVC. The ResNet-50 models with different GCP meta-layers are first pre-trained on ImageNet using the proposed hybrid training strategy and then fine-tuned on each FGVC dataset. Table~\ref{tab:fgvc_acc} compares their best and final validation accuracy. As can be observed, all the SVD variants have very competitive performances and surpass Newton-Schulz iteration by a large margin with 0.6\%, although some of them slightly trail Newton-Schulz iteration on ImageNet. This demonstrates our guess that accurate matrix square root also works better on FGVC datasets with small learning rate and well-trained network weights. Furthermore, the large margin may imply that the GCP networks trained by precise square root have better generalization capabilities than the approximate ones. Among these SVD variants, our proposed SVD-Pad\'e method achieves the best performance and outperforms Newton-Schulz iteration by 1\% across datasets. Besides, the SVD-TopN method cannot converge on any fine-grained datasets. This meets our expectation as the small eigenvalues may encode the class-specific features of the fine-grained classes and thus play an important role in FGVC. Despite the preliminary analysis, the problem is worth further investigation.

\subsection{Speed Comparison}

\begin{table}[htbp]
\vspace{-0.4cm} 
    \centering
    \caption{Computation time cost of each GCP meta-layer for a single batch on AlexNet and the computation bottleneck.}
     \resizebox{0.9\linewidth}{!}{
    \begin{tabular}{r|c|c|c|c}
    \toprule
    Methods  & Bottleneck & FP (s) & BP (s) & Total (s)\\
    \hline
    iSQRT-COV  & N/A &0.23 &0.53 &0.76 \\
    \hline \hline
    SVD-Pad\'e & \multirow{6}*{SVD} &0.96 &0.28 &1.24 \\
    SVD-Taylor & &0.96 &0.32 &1.28\\
    SVD-Newton & &2.36 &1.05 &3.41\\
    SVD-Trunc  & &0.97 &0.26 &1.23\\
    SVD-TopN   & &0.98 &0.24 &1.22\\
    MPN-COV    & &0.97 &0.23 &1.20\\
    \bottomrule
    \end{tabular}
    }
    \label{tab:speed}
\vspace{-0.2cm} 
\end{table}


We compare the time cost of a single batch for each meta-layer on AlexNet in Table~\ref{tab:speed}. Compared with iSQRT-COV~\cite{li2018towards}, our implementation consumes less back-propagation time as iterative matrix-matrix multiplication is not involved in the SVD backward algorithm. The computational bottleneck of our implementation is mainly the forward eigendecomposion, which is unfortunately limited by the platform. Our SVD-Pad\'e has faster backward speeds than SVD-Taylor, given that they agree up to the Taylor series of the same degree $K$. This is mainly because the numerator and denominator of Pad\'e approximants can be computed in parallel, the total iteration times would be $K{/}2$. For SVD-Taylor, it would take $K$ iterations to reach the same degree. 

\subsection{Approximation Error of Taylor and Pad\'e}

\begin{table}[htbp]
\vspace{-0.4cm} 
    \centering
    \caption{Approximation error of Taylor polynomial and diagonal Pad\'e approximants of degree 100 in double precision.}
     \resizebox{0.9\linewidth}{!}{
    \begin{tabular}{c|c|c|c|c|c|c|c}
    \toprule
    $\lambda_{j}/\lambda_{i}$  & 0.1 & 0.3 & 0.5 & 0.7 & 0.9 & 0.99 & 0.999\\
    \hline
    Taylor &9e-19 &7e-18 &2e-21 &8e-16 &2e-4 &36 &904\\
    Pad\'e &9e-19 &5e-18 &1e-21 &5e-17 &3e-16 &8e-13 &3e-10\\
    \bottomrule
    \end{tabular}
    }
    \label{tab:pade_vs_taylor}
\vspace{-0.3cm} 
\end{table}

Table.~\ref{tab:pade_vs_taylor} compares the approximation error to function $f(x){=}1{/}(1{-}x)$ of both Taylor polynomial and diagonal Pad\'e approximants of degree 100. As discussed in Sec.~\ref{sec:taylor_problem}, the approximation error of Taylor polynomial gets amplified when the eigenvalue ratio is close to the convergence boundary ($\lambda_{j}{/}\lambda_{i}{\approx}1$). By contrast, our Pad\'e approximants consistently provide good approximation for any $\lambda_{j}{/}\lambda_{i}$.

\subsection{Upper Bound of Gradient}

\begin{table}[htbp]
\vspace{-0.2cm} 
    \centering
    \setlength{\tabcolsep}{1.0pt}
    \caption{Upper bound of the gradient $K_{ij}$ for each SVD method.}
     \resizebox{0.9\linewidth}{!}{
    \begin{tabular}{r|c|c|c|c|c|c}
    \toprule
      Methods & SVD-Pad\'e & SVD-Taylor & SVD-Trunc & SVD-TopN & SVD-Newton &SVD\\
    \hline
      Analytical Form & $\frac{1}{\lambda_{i}}\cdot\frac{\sum\limits_{m=0}^{M}p_{m}}{1+\sum\limits_{n=1}^{N}q_{n}}$& $\frac{K+1}{\lambda_{i}}$ & $\rm{T}$ & $\frac{1}{\lambda_{N}}$ & /\ & $\frac{1}{\lambda_{i}-\lambda_{j}}$\\
      \hline
      Maximal Value &6.00e36 &4.55e17 &1e10 & 4.50e15 & /\ &$\infty$\\
      \hline
      Trigger Condition & $\lambda_{i}=\lambda_{j}\leq\textsc{eps}$ & $\lambda_{i}=\lambda_{j}\leq\textsc{eps}$  & $\Big| \frac{1}{\lambda_{i}-\lambda_{j}}\Big|\geq\rm{T}$& $\lambda_{N} \leq \textsc{eps} $ &/\ & $\lambda_{i}=\lambda_{j}$\\
    \bottomrule
    \end{tabular}
    }
    \label{tab:upper_bound}
\vspace{-0.3 cm} 
\end{table}

Table.~\ref{tab:upper_bound} summarizes the upper bound of gradient $K_{ij}$ for each SVD method. Our proposed SVD-Pad\'e allows for the largest upper bound of gradient, but the maximal value is still acceptable in the double precision (${<}1.79e308$) or even single precision (${<}3.40e38$). For the detailed derivation of upper bound, please refer to the Appendix.
\section{Conclusion}
In this paper, we empirically analyze why approximate matrix square root outperforms exact ones on GCP from data precision and gradient smoothness. We investigate various SVD remedies with smooth gradients and validate their performances on GCP. 
We suggest a hybrid training strategy to help the SVD methods to achieve competitive performances against Newton-Schulz iteration. Based on the findings, we propose a new GCP meta-layer that uses SVD as the forward pass and Pad\'e approximants during back-propagation for robust gradient approximation. The proposed meta-layer has achieved state-of-the-art performances on different datasets and deep models. 

{\small
\bibliographystyle{ieee_fullname}
\bibliography{egbib}
}

\appendix
\newpage
This document provides additional illustrations of the SVD functions. First, we analyze the convergence property of Power Iteration method and its impact on estimating the SVD gradient (Sec.~\ref{sec:0}). Then, we explain how the Taylor polynomial gradient emerges from the Power Iteration gradient~\cite{wang2019backpropagation} and their theoretical equivalence on certain premises (Sec.~\ref{sec:1}). Finally, several superior properties of Pad\'e approximants are introduced and proved (Sec.~\ref{sec:2}). 

\section{Convergence of Power Iteration}
\label{sec:0}
In the paper, we conjecture that Power Iteration (PI) method converges only when the first eigenvalue $\lambda_{1}$ is dominant. Here we analyze its convergence property and discuss the impact on the gradients associated with SVD~\cite{wang2019backpropagation}. To compute the approximate leading eigenvector $\mathbf{u}$, PI takes the iterative update:
\begin{equation}
    \mathbf{u}^{(k)}=\frac{\mathbf{P}\mathbf{u}^{(k-1)}}{||\mathbf{P}\mathbf{u}^{(k-1)}||}
    \label{pi}
\end{equation}
By induction on $k$, we have:
\begin{equation}
    \mathbf{u}^{(k)}=\frac{\mathbf{P}^{k}\mathbf{u}^{(0)}}{||\mathbf{P}^{k}\mathbf{u}^{(0)}||},\ k\geq 1
    \label{pi_final}
\end{equation}
Since $\mathbf{P}$ is diagonalizable, the initial eigenvector can be represented by a basis function of the true eigenvectors:
\begin{equation}
    \mathbf{u}^{(0)}=\sum\limits_{i=1}^{n}\alpha_{i} \mathbf{x}_{i}
    \label{u0}
\end{equation}
where $\alpha_{i}$ is the scalar coefficient, and $\mathbf{x}_{i}$ is the true eigenvector of $\mathbf{P}$. Injecting~\cref{u0} into~\cref{pi_final} yields:
\begin{equation}
    \mathbf{P}^{k}\mathbf{u}^{(0)}=\sum\limits_{i=1}^{n}\alpha_{i} \mathbf{P}^{k} \mathbf{x}_{i},\ k\geq1
    \label{Pk}
\end{equation}
Relying on the fact $\mathbf{P}\mathbf{x}_{i}{=}\lambda_{i}\mathbf{x}_{i}$,~\cref{Pk} can be re-formulated as:
\begin{equation}
    \mathbf{P}^{k}\mathbf{u}^{(0)}=\sum\limits_{i=1}^{n}\alpha_{i} \lambda_{i}^{k} \mathbf{x}_{i}=\alpha_{1}\lambda_{1}^{k}(\mathbf{x}_{1}+\sum\limits_{i=2}^{n}\frac{\alpha_{i}}{\alpha_{1}}(\frac{\lambda_{i}}{\lambda_{1}})^{k}\mathbf{x}_{i})
    \label{Pk2}
\end{equation}
If the first eigenvalue $\lambda_{1}$ is dominant, \emph{i.e.,} $\lambda_{i}$ satisfies the following condition:
\begin{equation}
    \lambda_{1} > \lambda_{2}\geq\lambda_{3}\geq\dots\geq\lambda_{n}
\end{equation}
When $k{\rightarrow}\infty$, for any $\frac{\lambda_{i}}{\lambda_{1}}$, $(\frac{\lambda_{i}}{\lambda_{1}})^{k}$ vanishes and ~\cref{Pk2} becomes $\mathbf{P}^{k}\mathbf{u}^{(0)}{\rightarrow}\alpha_{1}\lambda_{1}^{k}\mathbf{x}_{1}$. The constant $\alpha_{1}$ would be cancelled by the $l_{2}$ normalization of each step. Thus, the estimation $\mathbf{u}^{(k)}$ aligns itself to the direction of the first eigenvector $\mathbf{x}_{1}$ and the convergence is guaranteed.

When the first eigenvalue $\lambda_{1}$ is not dominant (\emph{i.e.,} $\lambda_{1}{=}\lambda_{2}$), $(\frac{\lambda_{2}}{\lambda_{1}})^{k}$ does not vanish and PI method cannot converge to the leading eigenvector $\mathbf{x}_{1}$. From~\cref{Pk2}, we can also learn that the convergence rate depends on the ratio of the first two eigenvalues $\frac{\lambda_{2}}{\lambda_{1}}$. The lower the ratio is, the faster the convergence would be. When $\frac{\lambda_{2}}{\lambda_{1}}$ is close to or equal to $1$ (see Fig. 3 in the paper), PI does not well approximate the leading eigenvector within limited iterations. As a consequence, the associated eigenvalue and gradients would be poorly estimated. 

\section{Relation between PI and Taylor Gradient}
\label{sec:1}
Wang~\emph{et. al.}~\cite{wang2019backpropagation} proposed to use PI method to compute the associated SVD gradients. They did not explicitly relate PI with the Taylor polynomial, but the relation between these two methods already emerged. To derive their connections, we first re-introduce the ordinary SVD gradients and PI gradient, then explain how Taylor polynomial gradient emerges from the former two methods, and end with the theoretical equivalence of PI and Taylor gradients on the premise that $\lambda_{1}$ is dominant.

\noindent \textbf{Ordinary SVD Gradient.}
Consider the covariance matrix $\mathbf{P}$, the forward eigendecomposition is given by:
\begin{equation}
    \mathbf{P}=\mathbf{U}\mathbf{\Lambda}\mathbf{U}^{T}
\end{equation}
where $\mathbf{U}$ and $\mathbf{V}$ are the corresponding eigenvector matrix and eigenvalue matrix, respectively. Given the loss function $l$, the partial derivative passed to $\mathbf{P}$ is computed as: 
\begin{equation}
    \frac{\partial l}{\partial \mathbf{P}}=\mathbf{U}( (\mathbf{K}^{T}\circ(\mathbf{U}^{T}\frac{\partial l}{\partial \mathbf{U}}))+ (\frac{\partial l}{\partial \mathbf{\Lambda}})_{\rm diag})\mathbf{U}^{T}
    \label{svd_derivative}
\end{equation}
where the skew-symmetric matrix $\mathbf{K}$ consists of elements $K_{ij}$ defined by:
\begin{equation}
    K_{ij}=\begin{cases}
    \frac{1}{\lambda_{i}-\lambda_{j}},\ i\neq j,\\
    0,\ \ \ \ \ \ \ \ \ i=j.
    \end{cases}
    \label{Kij}
\end{equation}
Injecting~\cref{Kij} into~\cref{svd_derivative} yields:
\begin{equation}
    \frac{\partial l}{\partial \mathbf{P}}=\sum\limits_{i=1}^{n}\sum\limits_{j\neq i}^{n}    \frac{1}{\lambda_{i}-\lambda_{j}}\mathbf{u}_{j}\mathbf{u}_{j}^{T}\frac{\partial l}{\partial \mathbf{u}_{i}} \mathbf{u}_{i}^{T} + \sum\limits_{i=1}^{n}\frac{\partial l}{\partial \lambda_{i}}\mathbf{u}_{i}\mathbf{u}_{i}^{T} 
    \label{svd_gradient}
\end{equation}
where $n$ is the total number of eigenvalues, $\mathbf{u}_{i}$ is the $i$-th row eigenvector of $\mathbf{U}$. The instability of the analytical gradient arises from
$\lambda_{i}{-}\lambda_{j}$ in the denominator of the first term. If the two eigenvalues are very close or even equal, the resultant gradient tends to be infinite and cause overflow. 

\noindent \textbf{Power Iteration Gradient.} As formulated in~\cite{ye2017dynamic}, the PI gradients can be computed as:
\begin{equation}
\begin{gathered}
     \frac{\partial l}{\partial \mathbf{P}}=\sum\limits_{k=0}^{K-1}\frac{\mathbf{I}-\mathbf{u}^{(k+1)}\mathbf{u}^{(k+1)T}}{||\mathbf{P}\mathbf{u}^{(k)}||} \frac{\partial l}{\partial \mathbf{u}^{(k+1)}} \mathbf{u}^{(k)T}, \\
     \frac{\partial l}{\partial \mathbf{u}^{(k)}}=\mathbf{P}\frac{\mathbf{I}-\mathbf{u}^{(k+1)}\mathbf{u}^{(k+1)T}}{||\mathbf{P}\mathbf{u}^{(k)}||}\frac{\partial l}{\partial \mathbf{u}^{(k+1)}}.
\end{gathered}
\label{pi_gradient}
\end{equation}
\cite{wang2019backpropagation} suggested that the initial eigenvector estimation starts with the accurate one calculated via SVD. Feeding it to Power Iteration defined in~\cref{pi} leads to:
\begin{equation}
    \mathbf{u} = \mathbf{u}^{(0)} \approx \mathbf{u}^{(1)} \approx \mathbf{u}^{(2)} \approx \dots \approx \mathbf{u}^{(K+1)}
\end{equation}
This equation generally holds when the first eigenvalue $\lambda_{1}$ is dominant. By exploiting this equation in~\cref{pi_gradient}, the gradient $\frac{\partial l}{\partial \mathbf{P}}$ can be re-written as:
\begin{equation}
\begin{split}
    \frac{\partial l}{\partial \mathbf{P}}= \Big( \frac{(\mathbf{I}-\mathbf{u}\mathbf{u}^{T})}{||\mathbf{P}\mathbf{u}||}+
    \frac{\mathbf{P}(\mathbf{I}-\mathbf{u}\mathbf{u}^{T})}{||\mathbf{P}\mathbf{u}||^{2}}
    +\\\cdots+
    \frac{\mathbf{P}^{K}(\mathbf{I}-\mathbf{u}\mathbf{u}^{T})}{||\mathbf{P}\mathbf{u}||^{K+1}} \Big)
    \frac{\partial l}{\partial \mathbf{u}} \mathbf{u}^{T}
    \label{pi_gradient1}
\end{split}
\end{equation}
Relying on
\begin{equation}
    \begin{gathered}
    \mathbf{P}^{k}=\lambda_{1}^{k}\mathbf{u}_{1}\mathbf{u}_{1}^{T} + \lambda_{2}^{k}\mathbf{u}_{2}\mathbf{u}_{2}^{T} +\cdots+ \lambda_{3}^{k}\mathbf{u}_{3}\mathbf{u}_{3}^{T},\\
    ||\mathbf{P}\mathbf{u}||^{k} = ||\lambda^{k}\mathbf{u}||^{k} = \lambda^{k}.
    \end{gathered}
\end{equation}
The gradient in~\cref{pi_gradient1} can be further formulated as:
\begin{equation}
   \begin{gathered}
       \frac{\partial l}{\partial \mathbf{P}}{=} \Big(\frac{\sum_{i{=}2}^{n}\mathbf{u}_{i}\mathbf{u}_{i}^{T}}{\lambda_{1}}{+} {\cdots}{+}
       \frac{\sum_{i{=}2}^{n}\lambda_{i}^{K}\mathbf{u}_{i}\mathbf{u}_{i}^{T}}{\lambda_{1}^{K+1}}\Big)\frac{\partial l}{\partial \mathbf{u}_{1}}\mathbf{u}_{1}^{T}, \\
       \frac{\partial l}{\partial \mathbf{P}}{=} \Big(\sum_{i=2}^{n}\Big( \frac{1}{\lambda_{1}}+\frac{1}{\lambda_{1}}(\frac{\lambda_{i}}{\lambda_{1}})^{1}+\cdots+\frac{1}{\lambda_{1}}(\frac{\lambda_{i}}{\lambda_{1}})^{K}\Big)\mathbf{u}_{i}\mathbf{u}_{i}^{T} \Big)\frac{\partial l}{\partial \mathbf{u}_{1}}\mathbf{u}_{1}^{T}\\
       \frac{\partial l}{\partial \mathbf{P}}{=}\Big(\sum_{i=2}^{n}\frac{1}{\lambda_{1}}\Big( 1+(\frac{\lambda_{i}}{\lambda_{1}})^{1}+\cdots+(\frac{\lambda_{i}}{\lambda_{1}})^{K}\Big)\mathbf{u}_{i}\mathbf{u}_{i}^{T} \Big)\frac{\partial l}{\partial \mathbf{u}_{1}}\mathbf{u}_{1}^{T}.
   \end{gathered}
   \label{geometric_progression}
\end{equation}
This equation defines a geometric progression. When $k\rightarrow\infty$, we have:
\begin{equation}
    \frac{1}{\lambda_{1}}\Big( 1+(\frac{\lambda_{i}}{\lambda_{1}})^{1}+\cdots+(\frac{\lambda_{i}}{\lambda_{1}})^{K}\Big) \approx \frac{1}{\lambda_{1}} \frac{1}{1-\frac{\lambda_{i}}{\lambda_{i}}} = \frac{1}{\lambda_{1}-\lambda_{i}}
    \label{taylor_geometric}
\end{equation}
If we read the equation from right to left, it is easy to find that~\cref{geometric_progression} actually defines the Maclaurin series of the Taylor expansion for function $\frac{1}{\lambda_{1}-\lambda_{i}}$. Consider the instability term $\frac{1}{\lambda_{i}-\lambda_{j}}$ of the SVD gradients in~\cref{svd_derivative}, the Taylor polynomial gradient naturally arises if we apply the same property on the ordinary SVD gradients.

\noindent \textbf{Taylor Polynomial Gradient.} To obtain the Taylor polynomial gradients, we first use the property in~\cref{taylor_geometric} to expand the skew-symmetric matrix $\mathbf{K}$ as:
\begin{equation}
    K_{ij}=\frac{1}{\lambda_{i}}\cdot\frac{1}{1-(\lambda_{j}/\lambda_{i})}\approx\frac{1}{\lambda_{i}}(1+\frac{\lambda_{j}}{\lambda_{i}}+(\frac{\lambda_{j}}{\lambda_{i}})^2+\dots+(\frac{\lambda_{j}}{\lambda_{i}})^K)
    \label{taylor_poly}
\end{equation}
Here it is the Taylor expansion of function $f(x){=}1{/}(1{-x})$ at $x{=}0$ to degree $K$, and the higher-order term is discarded. According to Cauchy root test, the Taylor series only converges when $|{\lambda_{j}}/{\lambda_{i}}|{<}1$. When ${j}{<}{i}$, the ratio ${\lambda_{j}}/{\lambda_{i}}$ is larger than $1$ and thus outsides the convergence radius. To avoid this issue, we can split the matrix $\mathbf{K}$ into two triangular parts and re-write the right first term of~\cref{svd_gradient} as:
\begin{equation}
    \sum_{i=1}^{n}\Big(\sum_{j>i}^{n}\frac{1}{\lambda_{i}-\lambda_{j}}\mathbf{u}_{j}\mathbf{u}_{j}^{T} \frac{\partial l}{\partial \mathbf{u}_{i}}\mathbf{u}_{i}^{T}-\sum_{j<i}^{n}\frac{1}{\lambda_{j}-\lambda_{i}}\mathbf{u}_{j}\mathbf{u}_{j}^{T} \frac{\partial l}{\partial \mathbf{u}_{i}}\mathbf{u}_{i}^{T}\Big)
    \label{K_ij_ji}
\end{equation}
where the first term is the upper triangle when $j{>}i$ and ${\lambda_{j}}/{\lambda_{i}}{\leq}1$, and the second term defines the lower triangle when $j{<}i$ and ${\lambda_{j}}/{\lambda_{i}}{\geq}1$. As $\mathbf{K}$ is skew-symmetric , we only need to calculate the upper part, \emph{i.e.,} the first term that can converge. Introducing the Taylor polynomial defined in~\cref{taylor_poly} into~\cref{K_ij_ji}, the first term is re-expressed as:
\begin{equation}
    \sum_{j>i}^{n}\frac{1}{\lambda_{i}}(1+\frac{\lambda_{j}}{\lambda_{i}}+(\frac{\lambda_{j}}{\lambda_{i}})^2+\dots+(\frac{\lambda_{j}}{\lambda_{i}})^K)\mathbf{u}_{j}\mathbf{u}_{j}^{T}\frac{\partial l}{\partial \mathbf{u}_{i}}\mathbf{u}_{i}^{T}
    \label{taylor_gradient}
\end{equation}
Notice that now the Taylor gradient defined in~\cref{taylor_gradient} is quite similar with the PI gradient in~\cref{geometric_progression}. If we set $i{=}1$ for~\cref{taylor_gradient}, which represents the derivative w.r.t. the dominant eigenvector $\mathbf{v}_{1}$, two equations are identical and Taylor polynomial gradient is equivalent to PI gradient. This remains the same also for the other eigenvectors. The equivalence holds when the first eigenvalue $\lambda_{1}$ is dominant and therefore PI gradient is valid. That being said, the Taylor polynomial gradient which emerges from the ordinary SVD gradient and PI gradient is actually a more general expression of the gradient calculated by PI method.

\section{Properties of Pad\'e Approximants}
\label{sec:2}
We introduce three relevant properties of Pad\'e approximants and give proofs on the first and the last theorems.

\noindent \textbf{Uniqueness of Solution.}
Pad\'e approximants are defined by matching a given Taylor series and have the unique solution for each matched pair. \cite{baker1964theory} gives the following theorem:

\begin{duplicate}
 Any existing ${[}{M}{/}{N}{]}$ Pad\'e approximants to their formal power series $A(x)$ has the unique solution.
\end{duplicate}

This theorem is easy to be proved. Suppose there are two such Pad\'e approximants $P(x){/}Q(x)$ and $U(x){/}V(x)$ of the same degree, they must satisfy $P(x){/}Q(x){-}U(x){/}V(x){=}O(x^{M{+}N{+}1})$ as both approximate the same series. Multiplying $Q(x)V(x)$ on both sides lead to $P(x){/}Q(x){=}U(x){/}V(x)$. Since by definition $Q(0){=}V(0){=}1$ and both $P$ and $Q$, $U$ and $V$ are relatively prime, we can conclude that the supposedly different approximants are the same.

\noindent \textbf{Special Case of Continued Fraction.}
Continued fraction is known as one of the best rational approximation techniques~\cite{powell1981approximation}. A general continued fraction expression takes the form as:
\begin{equation}
    C=b_{0}+\sum\limits_{i=1}^{\infty}\frac{a_{i}}{b_{i}+}=b_{0}+\frac{a_{1}}{b_{1}+\frac{a_{2}}{b_{2}+\frac{a_{3}}{b_{3}+\dots}}}
    \label{CF}
\end{equation}
where $a_{i}$ and $b_{i}$ are partial numerator and denominator, respectively. If the $n^{th}$ convergent of~\cref{CF} is denoted as a fraction ${A_{n}}{/}{B_{n}}$.
, the recursive relations of the successive convergents can be expressed as
\begin{equation}
\begin{gathered}
    A_{n+1}=b_{n+1}A_{n}+a_{n+1}A_{n-1},\\
    B_{n+1}=b_{n+1}B_{n}+a_{n+1}B_{n-1}.
    \label{rec_cf}
\end{gathered}
\end{equation}
Note that~\cref{rec_cf} resembles the recursive relation of numerator $P_{M}(x)$ and denominator $Q_{N}(x)$ of Pad\'e approximants. Therefore, Pad\'e approximants can be viewed as a special case of continued fraction. Particularly in diagonal case, we have: 
\begin{duplicate}
 The successive convergents of Jacobi-type continued fractions can be given by corresponding diagonal sequence $[1/0], [2/1], \dots$ of Pad\'e approximants.
\end{duplicate}
\noindent This theorem has been extensively proved in the literature~\cite{baker1964theory,baker1970pade,george1975essentials}. It allows us to associate Pad\'e approximants to continued fraction and calculate Pad\'e coefficients recursively using~\cref{rec_cf}. This kind of recursive computation is usually more stable than solving linear equations, as the solution of linear equations might be close to the singularities of Toeplitz matrices~\cite{kallrath2002rational}.

\noindent \textbf{Enlarged Convergence Range.} We present the following theorem in the paper without proof:
\begin{duplicate}
If the function $f(z)$ is a Stieltjes transform $f(z){=}\int_{a}^{b}\frac{1}{z-x}du(x)$ of a compactly supported measure $u(x)$ in $[a{,}b]$, then the associated $[N{+}1{/}N]$ diagonal Pad\'e approximants are orthogonal and there exists such function $r(x){>}1$ that the convergence $\lim\limits_{N\rightarrow\infty}|f(z){-}\frac{P_{N{+}1}(z)}{Q_{N}(z)}|^{\frac{1}{N}}{=}\frac{1}{r^2}$ is exponential in $[a{,}b]$. 
\end{duplicate}
The theorem describes the orthogonality constraints and convergence property of diagonal Pad\'e approximants. The orthogonality where $\int_{a}^{b}Q_{N}(x)P_{N{+}1}d\mu(x){=}0$ is not easy to be proved, the readers are kindly suggested to refer to~\cite{van2006pade} for a detailed review. Taking the orthogonality of $P_{N{+}1}$ as the condition in hand, we give a concise proof on the convergence. 
\begin{proof}
   Given that the function $f(z)$ is a Stieltjes transform $f(z){=}\int_{a}^{b}\frac{1}{z-x}d\mu(x)$ of a compactly supported measure $\mu(x)$ in $[a{,}b]$, the denominator polynomial in the Pad\'e approximation is orthogonal polynomial for the measure $\mu$ on the interval ${[}a{,}b{]}$:
   \begin{equation}
       \int_{a}^{b}x^{k}Q_{N}(x)d\mu(x)=0
       \label{q_orth}
   \end{equation}
   The numerator polynomial is given by
   \begin{equation}
       P_{N+1}(z)=\int_{a}^{b}\frac{Q_{N}(z)-Q_{N}(x)}{z-x} d \mu(x)
   \end{equation}
   By normalizing~\cref{q_orth}, the approximation error of Pad\'e approximants is calculated as:
   \begin{equation}
       f(z)-\frac{P_{N+1}(z)}{Q_{N}(z)}=\frac{1}{Q_{N}(z)}\int_{a}^{b}\frac{Q_{N}(x)}{z-x}d \mu(x)
   \end{equation}
   Observe that
   \begin{equation}
   \begin{gathered}
       Q_{N}(z)\int_{a}^{b}\frac{Q_{N}(x)}{z-x}d \mu(x)=
       \\\int_{a}^{b}\frac{Q_{N}(x)(Q_{N}(z)-Q_{N}(x))}{z-x}d\mu(x) +  \int_{a}^{b}\frac{Q_{N}^{2}(x)}{z-x}d \mu(x)
   \end{gathered}
   \end{equation}
   By orthogonality constraint the first integral on the right side vanishes. The error then becomes:
   \begin{equation}
       f(z)-\frac{P_{N+1}(z)}{Q_{N}(z)}=\frac{1}{Q_{N}^{2}(z)}\int_{a}^{b}\frac{Q_{N}^{2}(x)}{z-x}d \mu(x)
       \label{error}
   \end{equation}
   Now the error contains two parts $\frac{1}{Q_{N}^{2}(z)}$ and $\int_{a}^{b}\frac{Q_{N}^{2}(x)}{z-x}d \mu(x)$. The integral term is actually a Markov function for the probability measure $Q_{N}^{2}(x)d\mu(x)$ when $Q_{N}$ is orthonormal. It can be estimated by a strictly positive distance measure defined as:
   \begin{equation}
       d_{K}:=inf\{|z-x|:z\in K, x\in [a,b]\}
   \end{equation}
   where $K$ is a compact set that $z$ belongs to. Then we have:
   \begin{equation}
       \Bigg|\int_{a}^{b}\frac{Q_{N}^{2}(x)}{z-x}d \mu(x)\Bigg|\leq\int_{a}^{b}\frac{Q_{N}^{2}(x)}{|z-x|}d \mu(x)\leq\frac{1}{d_{K}}
   \end{equation}
   This bound is independent of the polynomial degree $N$. So the convergence is completely
    determined by the asymptotic behavior of $Q_{N}$. By measuring the logarithm energy for the leading coefficients of $Q_{N}$, \cite{van2006pade} further shows that
    \begin{equation}
        \lim\limits_{N\rightarrow\infty} |Q_{N}(z)|^{\frac{1}{N}}=\frac{4}{b-a}\exp{\Big(-\int_{a}^{b}\log\frac{1}{|z-x|}d\mu_{e}(x)\Big)}
        \label{asymptotic}
    \end{equation}
    where $\mu_{e}$ is a unique probability measure on $[a{,}b]$, and the right hand side is larger than $1$ when $z$ moves away from $[a{,}b]$. Let $r$ denotes the right hand side of~\cref{asymptotic}, the error in~\cref{error} can be re-formulated as:
    \begin{equation}
        \lim\limits_{N\rightarrow\infty}\Bigg|f(z)-\frac{P_{N+1}(z)}{Q_{N}(z)}\Bigg|=\frac{1}{r^2},\ r>1
    \end{equation}
    We can conclude that the diagonal Pad\'e approximants have exponential convergence in $[a{,}b]$.
\end{proof}

\newpage
This document introduces the experimental settings, some analyses on the SVD meta-layers, and extra ablation studies. First, we present the experimental settings in Sec.~\ref{sec:3}. Subsequently, Sec.~\ref{sec:4} and Sec.~\ref{sec:5} justify the degree selection of the Taylor series and compare the upper bound of the gradient for each SVD method, respectively. Finally, Sec.~\ref{sec:6} describes the results of ablation studies on the random seeds and warm-up epochs.

\section{Experimental Settings}
\label{sec:3}

\noindent \textbf{Models and Datasets.}
Following~\cite{li2017second,li2018towards}, we first take AlexNet~\cite{krizhevsky2017imagenet} and ResNet-50~\cite{he2016deep} as the backbones and conduct experiments on ImageNet 2012~\cite{deng2009imagenet} for the large-scale visual recognition. This dataset has $1.28$M images for training and $50$K images for testing. The covariance pooling meta-layer is inserted before the fully-connected layer of each model. For AlexNet architecture, the outputs of convolutional layers are $13{\times}13{\times}256$ tensor. For ResNet architecture, we add $1{\times}1$ convolution to squeeze the channels of global representation from $2048$ to $256$. Therefore, the covariance matrices of both networks are of the same size $256{\times}256$. Since the covariance is a symmetric matrix, only the upper triangular part is taken and passed to the fully-connected layer. After training GCP models on ImageNet, we then conduct experiments on the task of Fine-Grained Visual Categorization (FGVC). The pre-trained ResNet-50 with different GCP meta-layers using the hybrid training strategy are fine-tuned on three popular fine-grained benchmarks, \emph{i.e.,} Caltech Birds (Birds)~\cite{WelinderEtal2010}, Stanford Dogs (Dogs)~\cite{KhoslaYaoJayadevaprakashFeiFei_FGVC2011}, and Stanford Cars (Cars)~\cite{KrauseStarkDengFei-Fei_3DRR2013}. The Birds dataset contains $11,788$ images belonging to $200$ species. The Dogs dataset includes $20,580$ images of $120$ breeds of dogs, and the Cars dataset consists of $16,185$ images from $196$ classes of cars.

\noindent \textbf{Implementation Details.}
All the source codes are implemented in Pytorch. Except that the forward eigendecomposition is performed on CPU for faster speed, the other operations are conducted on GPU. For the SVD-TopN method, we keep the top $200$ out of $256$ eigenvalues. The maximum gradient for the SVD-Trunc method is limited to $10^{10}$. We set the iteration times as $10$ for the SVD-Newton method. For SVD-Taylor and SVD-Pad\'e, we truncate the Taylor series to degree 100 and match the diagonal Pad\'e approximants also to degree 100, respectively. During the forward pass, the eigenvalues that are smaller than \textsc{eps}, \emph{i.e.,} the smallest positive number that the data precision can represent, are set as \textsc{eps} for numerical stability. 


\noindent \textbf{ImageNet Setting.}
On ImageNet, the AlexNet is trained for 30 epochs with an initial learning rate set as $10^{-1.1}$. The learning rate decays by $10$ every $10$ epochs. For training ResNet, we use the same learning rate to train for $60$ epochs but decays by $10$ at epoch $30$ and epoch $45$. When applying the hybrid training strategy on AlexNet, all the methods are warmed up for 1 epoch when switching to SVD methods. The warm-up epoch for ResNet is set as 2. The batch size is set to $128$ for AlexNet and $256$ for ResNet. We use SGD for optimization, with momentum of $0.9$ and weight decay of $0.0001$ for ResNet and $0.0005$ for AlexNet. The network parameters are randomly initialized for both AlexNet and ResNet. During training, the images are resized to $256{\times}256$ and then cropped to $224{\times}224$, with random horizontal flip augmentation. The inference is conducted on the $224{\times}224$ centered crop from the test image. 

\noindent \textbf{FGVC Setting.}
For FGVC datasets, the images are first resized to $448{\times}448$ and then fed into the network. The $1000$-$d$ fully-connected layer of the original model is changed to fit the number of classes. The model is trained using SGD with momentum $0.9$ and weight decay $0.0001$. The batch size is set 10, and the training lasts $50$ epochs for all the datasets. The learning rate is set as $6{\times}10^{-3}$ for the fully-connected layer and $1.2{\times}10^{-3}$ for the other layers. We make the inference on the $448{\times}448$ centered crop of the test image.

\section{Choosing Degree of Taylor Series}
\label{sec:4}
Both SVD-Taylor and SVD-Pad\'e need to match the truncated Taylor series of degree $K$. For SVD-Taylor, $K$ determines the upper bound of gradient approximation ${(K{+}1)}{/}{\lambda_{i}}$ and the discarded higher-order term $\sum_{i{=}K{+}1}^{\infty} (\lambda_{j}{/}\lambda_{i})^i$. As can be observed from Table~\ref{tab:taylor_degree}, a small $K$ will yield poor approximation, but a large $K$ increases the computation time and still fails to well approximate values close to the convergence boundary ($\lambda_{j}/\lambda_{i}\approx 1$). Unless $K$ is set very large (\emph{e.g.,} $10^{20}$), gradients near the polar singularities can not closely estimated. Thus, choosing an appropriate $K$ can substantially influence the gradient approximation. For SVD-Pad\'e, $K$ has a slight effect on the approximation due to the enlarged convergence range of Pad\'e approximants (see Table~\ref{tab:pade_degree}). The degree $K$ is set as $100$ through cross-validation for the best performances of SVD-Taylor.  We also choose $K{=}100$ for SVD-Pad\'e to make sure that both methods agree up to the same degree.

\begin{table}[htbp]
    \centering
    \caption{Approximation error of Taylor polynomial of different degrees in double precision.}
     \resizebox{0.99\linewidth}{!}{
    \begin{tabular}{c|c|c|c|c|c|c|c}
    \toprule
    \diagbox{Degree}{$\lambda_{j}/\lambda_{i}$}   & 0.1 & 0.3 & 0.5 & 0.7 & 0.9 & 0.99 & 0.999\\
    \hline
    50 &9e-19 &7e-18 &9e-16 &4e-8 &5e-2 &60 &950\\
    100 &9e-19 &7e-18 &2e-21 &8e-16 &2e-4 &36 &904\\
    200 &9e-19 &7e-18 &2e-21 &4e-17 &6e-9 &13 &817\\
    300 &9e-19 &7e-18 &1e-21 &4e-17 &1e-13 &5 &740\\
    \bottomrule
    \end{tabular}
    }
    \label{tab:taylor_degree}
\end{table}

\begin{table}[htbp]
    \centering
    \caption{Approximation error of diagonal Pad\'e approximants of different degrees in double precision.}
     \resizebox{0.99\linewidth}{!}{
    \begin{tabular}{c|c|c|c|c|c|c|c}
    \toprule
    \diagbox{Degree}{$\lambda_{j}/\lambda_{i}$}  & 0.1 & 0.3 & 0.5 & 0.7 & 0.9 & 0.99 & 0.999\\
    \hline
    50  &2e-18 &3e-17 &2e-20 &6e-17 &3e-16 &1e-13 &1e-12\\
    100 &9e-19 &5e-18 &1e-21 &5e-17 &3e-16 &8e-13 &3e-10\\
    200 &2e-18 &1e-17 &6e-21 &6e-18 &1e-15 &1e-13 &2e-10\\
    300 &1e-18 &8e-18 &6e-21 &5e-17 &2e-15 &1e-13 &5e-10\\
    \bottomrule
    \end{tabular}
    }
    \label{tab:pade_degree}
\end{table}

\section{Upper Bound of Gradient}
\label{sec:5}
We first describe how the upper bound is attained for each method in detail and then discuss their behaviors.

\noindent \textbf{SVD-Pad\'e.}  Both SVD-Pad\'e and SVD-Taylor decompose the gradient function as:
\begin{equation}
    K_{ij}=\frac{1}{\lambda_{i}-\lambda_{j}}=\frac{1}{\lambda_{i}}\frac{1}{1-\lambda_{j}/\lambda_{i}}
    \label{gradient_compostion}
\end{equation}
The term $\frac{1}{1-\lambda_{j}/\lambda_{i}}$ can be viewed as the function $f(x){=}\frac{1}{1-x}$ and we use Pad\'e approximants to approximate it. Since the function $f(x)$ is monotonically increasing in the range $[0{,}1]$, the upper bound of Pad\'e approximants is reached at $x{=}1$, \emph{i.e.,} when the two eigenvalues $\lambda_{i}$ and $\lambda_{j}$ are identical. The upper bound for \cref{gradient_compostion} can be represented as:
\begin{equation}
    |K_{ij}|\leq\Big|\frac{1}{\lambda_{i}}\Big|\cdot\Big|\frac{\sum\limits_{m=0}^{M}p_{m}}{1+\sum\limits_{n=1}^{N}q_{n}}\Big|
    \label{pade_bound}
\end{equation}
The second fraction on the right side denotes the maximal value of Pad\'e approximants when $\frac{\lambda_{j}}{\lambda_{i}}{=}1$. We compute this result as $6.48e20$. Now the upper bound of gradient depends on $\frac{1}{\lambda_{i}}$. Suppose $\lambda_{i}$ and $\lambda_{j}$ simultaneously equal to \textsc{EPS}, the resultant upper bound of SVD-Pad\'e is attained. The bound is calculated as $6.00e36$ and it happens only when the two eigenvalues simultaneously have the minimum possible value ($\lambda_{i}{=}\lambda_{j}{=}\textsc{eps}$). 

\noindent \textbf{SVD-Taylor.} For SVD-Taylor, the upper bound relies on the truncated degree of Taylor series $K$ and the minimum value of $\lambda_{i}$. Specifically, the Taylor polynomial is a bounded estimation:
\begin{equation}
    \frac{1}{1-\lambda_{j}/\lambda_{i}}\approx1+\frac{\lambda_{j}}{\lambda_{i}}+(\frac{\lambda_{j}}{\lambda_{i}})^2+\dots+(\frac{\lambda_{j}}{\lambda_{i}})^K\leq K+1
    \label{taylor_K_series}
\end{equation}
The equality is taken if $\lambda_{i}{=}\lambda_{j}$. Combining~\cref{taylor_K_series} with~\cref{gradient_compostion}, the analytical form of the upper bound can be derived as: 
\begin{equation}
    |K_{ij}|\leq \Big|\frac{K+1}{\lambda_{i}}\Big|
\end{equation}
Similar with SVD-Pad\'e, when $\lambda_{i}{=}\lambda_{j}{=}\textsc{eps}$, the upper bound is attained as $4.55e17$.

\noindent \textbf{SVD-Trunc.} As we directly truncate the gradient $K_{ij}$ by a large constant $\rm{T}$ for SVD-Trunc, the upper bound of gradient is equal to $\rm{T}$. The truncation and also the upper bound are triggered when $\Big|\frac{1}{\lambda_{i}-\lambda_{j}}\Big|{\geq}\rm{T}$.

\noindent \textbf{SVD-TopN.} For SVD-TopN, the upper bound of gradient is very likely to happen between the last kept eigenvalue $\lambda_{N}$ and the first abandoned eigenvalue $\lambda_{N{+}1}$. As $\lambda_{N{+}1}$ is truncated to zero, the bound takes the form $\frac{1}{\lambda_{N}}$. The maximal value is reached when $\lambda_{N}{=}\textsc{eps}$.

\noindent \textbf{SVD-Newton.} As the iterative matrix-matrix product is involved in the backward algorithm of Newton-Schulz iteration, the upper bound for the SVD-Newton method can not be derived. But from the empirical observation on the effective $\beta$-smoothness~\cite{nesterov2003introductory} (see Fig. 4 right in the paper), the gradient is very smooth and the upper bound is expected to be similar with SVD-Trunc. 

\noindent \textbf{SVD.} The ordinary SVD gradient takes the form $\frac{1}{\lambda_{i}{-}\lambda_{j}}$. When the two eigenvalues are equal, the gradient will explode and go to infinity.

Table.~\ref{tab:upper_bound_full} summarizes the upper bound of gradient $K_{ij}$ for each SVD variant and their happening conditions. Compared with the ordinary SVD gradients, these SVD remedies reduce both the magnitude and the occurrence of the upper bound.  Our proposed SVD-Pad\'e allows for the largest gradient upper bound, but the maximal value is still acceptable in the double precision (${<}1.79e308$). Even for the single precision (${<}3.40e38$), the gradient is also numerically stable and allowed. This can ensure that the SVD-Pad\'e meta-layer is compatible with the backbone either in single or double precision. The compatibility also shows the possibility that our SVD meta-layers can be trained by the recent advanced mixed-precision training techniques (\emph{e.g.,} Pytorch 1.8 and Nvidia Apex 1.0) for acceleration and stability.


\begin{table}[htbp]
    \centering
    \caption{Upper bound of the gradient $K_{ij}$ for each SVD method.}
     \resizebox{0.99\linewidth}{!}{
    \begin{tabular}{r|c|c|c|c|c|c}
    \toprule
      Methods & SVD-Pad\'e & SVD-Taylor & SVD-Trunc & SVD-TopN & SVD-Newton &SVD\\
    \hline
      Analytical Form & $\frac{1}{\lambda_{i}}\cdot\frac{\sum\limits_{m=0}^{M}p_{m}}{1+\sum\limits_{n=1}^{N}q_{n}}$& $\frac{K+1}{\lambda_{i}}$ & $\rm{T}$ & $\frac{1}{\lambda_{N}}$ & /\ & $\frac{1}{\lambda_{i}-\lambda_{j}}$\\
      \hline
      Maximal Value &6.00e36 &4.55e17 &1e10 & 4.50e15 & /\ &$\infty$\\
      \hline
      Trigger Condition & $\lambda_{i}=\lambda_{j}\leq\textsc{eps}$ & $\lambda_{i}=\lambda_{j}\leq\textsc{eps}$  & $\Big| \frac{1}{\lambda_{i}-\lambda_{j}}\Big|\geq\rm{T}$& $\lambda_{N} \leq \textsc{eps} $ &/\ & $\lambda_{i}=\lambda_{j}$\\
    \bottomrule
    \end{tabular}
    }
    \label{tab:upper_bound_full}
\end{table}

\section{Ablation Studies}
\label{sec:6}
\noindent \textbf{Impact of Random Seed.}
We measure the impact of random seeds by having 5 runs for SVD-Pad\'e meta-layer on AlexNet using the standalone training strategy. Different random seeds do influence the network in the early epochs, but the impact gets weakened in the later stage. The final error fluctuates within $0.1\%$. This variation would not shake our deductions, as our SVD-Pad\'e meta-layer outperforms the other SVD remedies by at least $0.2\%$. We expect that the fluctuation would be similar or smaller for ResNet and hybrid training strategy.

\noindent \textbf{Impact of Warm-up Epochs.}
We take our proposed SVD-Pad\'e as the meta-layer and evaluate the impact of warm-up epochs when using the hybrid training strategy. As can be seen from Table~\ref{tab:alexnet_warm_top15}, two epochs achieve the best results in the final error. If warming up for more epochs, no obvious performance gain is observed in the final error but the best error continues to improve. It is also worth mentioning that the performance is still competitive even without any warm-up training. We set to 1 epoch for AlexNet in order not to introduce heavy burdens on the training process.  

\begin{table}[htbp]
    \centering
    \caption{Validation error of AlexNet using SVD-Pad\'e meta-layer and hybrid the training strategy with various warm-up epochs. The best three results are highlighted in \textcolor{red}{red}, \textcolor{blue}{blue}, and \textcolor{green}{green}.}
     \resizebox{0.99\linewidth}{!}{
    \begin{tabular}{r|cc|cc}
    \toprule
    \multirow{2}*{Settings} & \multicolumn{2}{c|}{Final Error (\%)} & \multicolumn{2}{c}{Best error (\%)}\\
   \cline{2-5}
   &top-1 & top-5 &top-1 & top-5 \\
    \hline
    no warm-up                     &\textcolor{green}{47.89} & 23.82 &47.75 &23.63\\
    1 epoch                        &\textcolor{blue}{47.76} & \textcolor{blue}{23.48} &\textcolor{green}{47.63} & \textcolor{blue}{23.21}\\
    2 epochs                       &\textcolor{red}{47.70} & \textcolor{red}{23.39}&\textcolor{blue}{47.59} & \textcolor{green}{23.23}\\
    3 epochs                       &47.95 & \textcolor{green}{23.57} &\textcolor{red}{47.54} &\textcolor{red}{23.14}\\
    \hline
     iSQRT-COV~\cite{li2018towards} &47.95 & 23.64 & 47.81 & 23.54\\
     \bottomrule
    \end{tabular}
    }
    \label{tab:alexnet_warm_top15}
\end{table}

\end{document}